\documentclass{article}

\usepackage{PRIMEarxiv}

\usepackage[utf8]{inputenc} 
\usepackage[T1]{fontenc}    
\usepackage{hyperref}       
\usepackage{url}            
\usepackage{booktabs}       
\usepackage{amsfonts}       
\usepackage{nicefrac}       
\usepackage{microtype}      
\usepackage{lipsum}
\usepackage{fancyhdr}       
\usepackage{graphicx}       
\graphicspath{{media/}}     

\usepackage{listings}
\usepackage{longtable}
\usepackage{multirow}
\usepackage{amsfonts}
\usepackage{amsmath}
\usepackage{booktabs} 
\usepackage{array}
\newcolumntype{L}[1]{>{\raggedright\let\newline\\\arraybackslash\hspace{0pt}}m{#1}}
\newcolumntype{C}[1]{>{\centering\let\newline\\\arraybackslash\hspace{0pt}}m{#1}}
\newcolumntype{R}[1]{>{\raggedleft\let\newline\\\arraybackslash\hspace{0pt}}m{#1}}
\usepackage{algorithm}
\usepackage{algpseudocode}
\usepackage{caption}
\usepackage{subcaption}
\usepackage{float}

\title{Optimisation of Federated Learning Settings under Statistical Heterogeneity Variations}

\author{
  Basem Suleiman \\
  University of New South Wales \\
  Australia\\
  \texttt{b.suleiman@unsw.edu.au} \\
   \And
  Muhammad Johan Alibasa \\
  Sampoerna University \\
  Indonesia\\
  \And
  Rizka Widyarini Purwanto \\
  Monash University \\
  Indonesia \\
  \texttt{rizka.purwanto@monash.edu} \\
  \AND
  Lewis Jeffries \\
  University of Sydney \\
  Australia \\
\And
  Ali Anaissi \\
  University of Sydney \\
  Australia \\
  \And
  Jacky Song \\
  University of Sydney \\
  Australia \\
}

\begin{document}
\maketitle

\begin{abstract}
Federated Learning (FL) enables local devices to collaboratively learn a shared predictive model by only periodically sharing model parameters with a central aggregator.
However, FL can be disadvantaged by statistical heterogeneity produced by the diversity in each local device's data distribution, which creates different levels of Independent and Identically Distributed (IID) data. Furthermore, this can be more complex when optimising different combinations of FL parameters and choosing optimal aggregation. In this paper, we present an empirical analysis of different FL training parameters and aggregators over various levels of statistical heterogeneity on three datasets. We propose a systematic data partition strategy to simulate different levels of statistical heterogeneity and a metric to measure the level of IID. Additionally, we empirically identify the best FL model and key parameters for datasets of different characteristics. On the basis of these, we present recommended guidelines for FL parameters and aggregators to optimise model performance under different levels of IID and with different datasets. 
\end{abstract}



\section{Introduction}\label{sec1}
Advancements in Artificial Intelligence (AI) have enabled increasingly personalised smart services. With huge amounts of data being collected in real-time throughout all areas of our lifestyle, service providers across all industries are rushing to equip themselves with the knowledge and methodologies to understand our ever-changing demands and behaviours. 
For example, machine learning (ML) algorithms can learn from connected wearable devices to ensure that patients can benefit from a more personalised and data-driven standard of care from a healthcare professional \cite{xu2021federated, beniczky2021machine}. Such advancements pose considerations regarding both data privacy amongst the user base and learning efficiency for data scientists.

Traditional machine learning algorithms are centralised in their approach. 
For example, all local data is sent to a common server which facilitates the model training centrally, before releasing a global update back to all devices. 
However, as more devices and sensors are connected, this raises the issue of privacy and security concerns because such devices can potentially collect highly sensitive user information \cite{abdulrahman2020survey, mothukuri2021survey}. Furthermore, it is expensive and inefficient to facilitate communication and store the increasingly collected and generated data throughout the network \cite{patra2016improving}. 
A Federated Learning (FL) approach has recently been introduced to address such concerns. FL decentralises the model training so that the participating devices all train on their own local models and only share their respective model updates to a central server, thus preserving raw data locally and reducing communication load \cite{zhu2021federated}. 
The central server subsequently aggregates all the federated model parameters to push the next global update to local devices, and the process repeats itself until a stop criterion is met. Studies comparing Centralised and Federated Learning across a variety of standard datasets have concluded that not only can FL achieve a higher test accuracy of approximately 25\%, it can also match the performance of a centralised model by engaging significantly less devices at each communication round \cite{asad2021federated}. 

The traditional approach to FL takes an average of the local model parameters using the FedAvg method \cite{mcmahan2017communication}. However, although this method yields highly favourable results in a theoretical setting, it is known to suffer from system and statistical heterogeneity upon scaling \cite{li2020federated}. Statistical heterogeneity indicates a scenario in which data are not independent and identically distributed (IID) between devices. Research \cite{zhao2018federated} shows that when using the FedAvg algorithm in one-class low-IID dataset, the test accuracy reduces by 55\% compared to a high-IID dataset. This is an important consideration, as the FL approach takes away the ability to address statistical heterogeneity across all devices centrally and instead relies on individual devices to train on their own local datasets, which can realistically inherit a certain degree of skewness. 

The goal of this research is to empirically find the optimal model using four FL aggregators - FedAvg, FedProx, FedPer, and SCAFFOLD - under various levels of statistical heterogeneity, which is determined by label distribution skewness and data quantity skewness. Our contributions are fourfold. First, we empirically analyse different FL aggregators and training parameters over various levels of statistical heterogeneity using three datasets (MNIST, CIFAR-10, and Health Care Obesity). Second, we propose a novel approach to a systematic data partition strategy to simulate different levels of statistical heterogeneity in FL settings. Third, we empirically identify the best FL model and key associated parameters, e.g., number of clients, number of communication rounds, and the number of samples needed in each local model. Forth, based on our empirical evaluation and analysis, we present guidelines for recommended FL settings given different levels of statistical heterogeneity and data characteristics. This can guide researchers in determining the optimal FL parameters and aggregators given the characteristics of their datasets. 




\section{Related Work}



One of the issues in federated learning is when the training data on local devices are not independent and identically distributed (non-IID) \cite{zhu2021federatedsurvey}. This issue leads to a worse performance of the federated learning model compared to the models that are trained using a non-federated learning (centralised) approach due to inconsistent update optimisation across every devices \cite{ye2023heterogeneous}. Several past studies discussed federated learning approaches and model architectures that are robust on heterogeneous data. In terms of model architecture, Qu et al. \cite{qu2022rethinking} found that self-attention-based models (e.g., Transformers) are suitable for deferated learning over heterogeneous data due to its robustness to distribution shifts. On the other hand, our study focuses on the optimal aggregation technique, which could be applied to federated learning in general regardless of the selected model architecture.

There are various methods to quantify and measure the effect of statistical heterogeneity on overall model performance \cite{li2020federated}. There is no one-size-fits-all approach to measuring this variable; however, we reviewed three different approaches to achieve this goal.
The Earth Mover's Distance (EMD) approach has been used to measure the degree of local data class imbalance in FL scenarios \cite{zhao2018federated,phan2022deepface}. This methodology quantifies the degree of skewness of the data label by measuring the variation in the distribution of data classes across the local datasets. The outcome of this study revealed a positive correlation between the EMD and the resulting weight divergence between localised and central models. Another study \cite{ma2021client} improved this approach by proposing a grouping-based scheduling method to select the client that ensures that the label distributions are balanced. The study found that this method improved the performance results.

Furthermore, studies have also used the Dirichlet distribution as a measure of label distribution in local datasets \cite{hsu2019measuring,vahidian2023rethinking}. This approach generates local data of various skewnesses by taking a vector of data label probabilities sampled from the Dirichlet distribution. This led to the conclusion that as the distribution of the labels varied increasingly between local devices, this contributed to a decrease in the overall performance of the model.
Finally, a Quantity-based Label Distribution Skew \cite{li2021federated} is also an effective method to simulate different degrees of local data skewness. This approach randomly assigns k different labels to each local device and ensures that there are no repeats of samples across the federated network. This study found that the number of local data labels (k) had a significant impact on model accuracy, especially comparing one-class low-IID settings to other levels of IID settings. 

Many advances in traditional FedAvg have been proposed, one of which is FedProx \cite{li2020federatedB}. There are two particular challenges of FedAvg that FedProx attempts to overcome: systems and statistical heterogeneity. First, the problem of system heterogeneity refers to the degree to which participating devices vary in terms of their capacity to train and share models with factors such as memory, CPU power, network connectivity, and battery life. Some studies make the assumption that there is full participation of all devices in the network; however, for almost all practical applications of FL, this is unrealistic due to data transfer bottlenecks \cite{qin2021federated}. Furthermore, variations in device memory and capacity will restrict the degree of meaningful contribution of learnings from devices to the central model. Taken together, some devices will be able to participate more fully than others. One of the contributions of FedProx is the ability to tolerate partial work of a device, whereby updates are allowed if they meet some threshold of improvement \cite{li2020federatedB}. Another significant contribution from FedProx is the addition of a proximal (i.e., regularisation) term. The purpose of the proximal term is to address the problem of statistical heterogeneity. The proximal term, like other regularisation methods in machine learning, seeks to penalise large weight updates to help keep the local updates from diverging; particularly in the case where a local model may be more affected by statistical heterogeneity. 

Stochastic Controlled Averaging for Federated Learning (SCAFFOLD) is another variation in the FL aggregator that has been shown to outperform FedAvg on image datasets in low-IID scenarios \cite{li2021federated}. SCAFFOLD produces a similar result to the regularisation method discussed above with FedProx, as it is designed to reduce the variance of changes in local model updates by introducing a control variate, helping to achieve faster convergence in the overall accuracy score relative to FedProx and FedAvg \cite{karimireddy2019scaffold}.
Recently, \cite{zhang2022federated} proposed Federated learning via Logits Calibration (FedLC) that showed improved performance in highly skewed data. This approach reduces significantly the possibility of overfitting to minority classes and missing classes by performing calibration of the logits of before softmax cross-entropy. The approach also included pairwise label margin during this calibration process of each local update.

FL algorithms have also addressed the problem of statistical heterogeneity with methods other than common machine learning parameter-based approaches such as regularisation or momentum. One such approach is that of FedPer \cite{arivazhagan2019federated}, which takes advantage of the federated environment and its many clients by adding personalisation layers to each of the local neural network models. The idea behind this contribution is that the data exposed to each local model will vary due to statistical heterogeneity, and this should ideally be addressed by local models at the device level. When a personalised learnable layer is added to the base model on each local server, the local model can learn better from its own data, rather than when every layer of the model is always updated with a globally aggregated layer. Consequently, the addition of the personalised layer(s) increases the effectiveness of the base model that is common to all the servers in the network. Once local training is completed in each round, the base model parameters (excluding the personalised layer parameters) are shared and aggregated as in FedAvg. Compared to FedAvg, FedPer has a faster convergence speed in terms of global communication rounds, as well as lower variance in test accuracies and higher test accuracy overall, especially when trained on a relatively low IID dataset.


\section{Methodologies}


To simulate real-world FL scenarios, where devices have their own local data in isolation from each other, with various classes distributions to simulate personal usage behaviour, we will introduce label distribution skewness and data quantity skewness \cite{li2021federated} to the local data of each device. Label distribution skewness refers to the imbalanced or uneven allocation of labels to various local clients in the federated learning system. Meanwhile, data quantity skewness refers to the variation in the number of data samples assigned to each local device in the federated learning system. The skewness in label distribution and data quantity on each local dataset can impact the performance of federated learning systems. In federated learning systems, model updates are aggregated from local devices. Label distribution skewness and data quantity skewness, also known as statistical heterogeneity, may lead to biased or uneven contributions from different devices, resulting in a biased global parameter. Furthermore, statistical heterogeneity influences convergence behaviour and the time required to reach a global consensus in the federated learning process \cite{li2020federated}.

\subsection{Data Collection}

In this study, three different datasets were used: MNIST, CIFAR-10 and Health Care Obesity dataset. The MNIST (Modified National Institute of Standards and Technology) dataset of handwritten digits was downloaded directly from the official website\footnote{{http://yann.lecun.com/exdb/mnist/}}. It has a training set of 60,000 examples and a test set of 10,000 examples. The digits have been size-normalised and centred in a fixed-size image. The CIFAR-10 dataset sourced from the official website\footnote{{https://www.cs.toronto.edu/~kriz/cifar.html}} consists of 60,000 32x32 colour images in 10 classes, with 6000 images per class. There are 50,000 training images and 10,000 test images.

\begin{figure}
\centering
\includegraphics[width=1\textwidth]{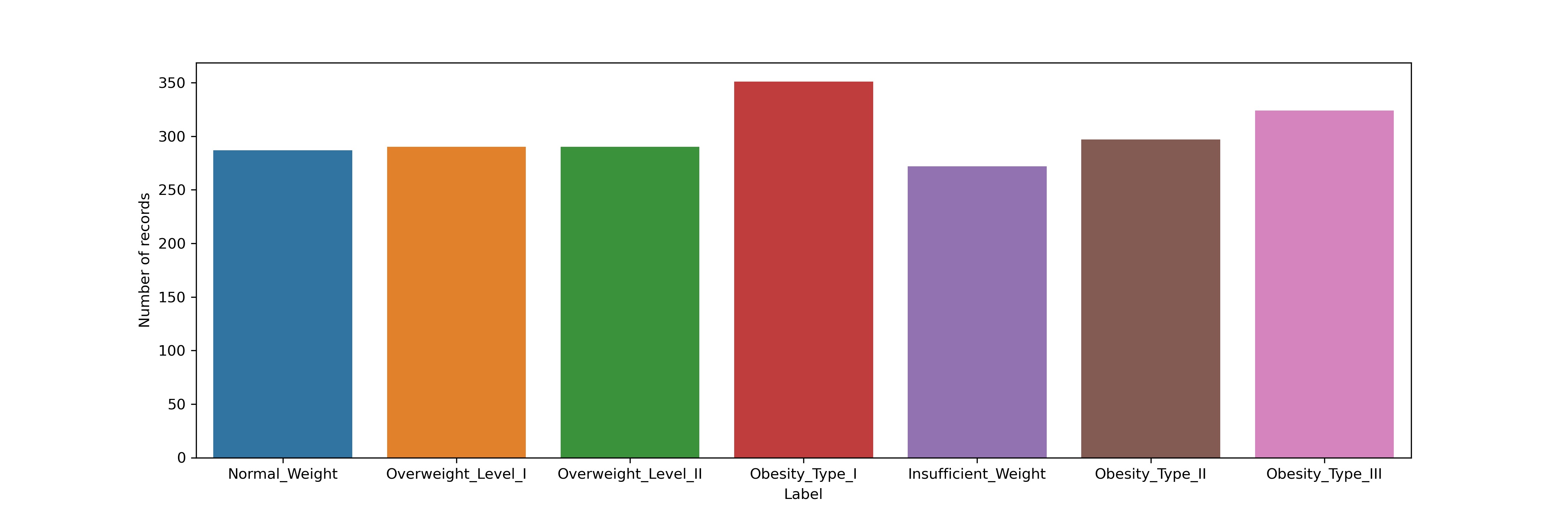}
\caption{Label distribution in the Health Care Obesity dataset}
\label{fig:dist_labels_obesity}
\end{figure}

The Health Care Obesity dataset was sourced from Kaggle\footnote{{https://www.kaggle.com/mpwolke/obesity-levels-life-style/notebook} Version 1}. It presents data for the estimation of an individual's levels of obesity based on their eating habits and physical condition \cite{palechor2019dataset}. It has a total of 2,111 rows, 16 variables, and 7 target labels, namely Insufficient Weight, Normal Weight, Overweight Level I, Overweight Level II, Obesity Type I, Obesity Type II, and Obesity Type III. Data balancing techniques has been applied to the dataset, resulting in a balanced distribution of labels across the dataset as shown in the Fig.~\ref{fig:dist_labels_obesity} that illustrates the balanced distribution of labels, which includes seven target labels. 

\subsection{Data Partitioning}\label{section:methods_datapartition}

One of the main contributions of our paper is to analyse FL aggregators where devices have varying distributions of data. Specifically, we are interested in the degree to which the data on the local devices can be made more or less IID to test aggregators under different conditions of skewness. As such, we put forward a novel method, different from those found in the literature, to quantify the extent to which data vary amongst the network of devices. 

There are two mechanisms that we consider as ways to vary the skewness of data assigned to each client: label skew and data quantity skew. Label skew refers to the number of distinct labels assigned to each active device. A smaller number of labels represents a lower IID setting, whereas a larger number of labels represents a higher IID setting. Devices may also be subjected to data quantity skew. When each device in the system has a less similar number of data samples to other devices in the federated system, that is, that there is increased data quantity variance amongst the devices, it can be considered as a lower IID setting. Likewise, when each device has a more similar number of samples (i.e., lower data quantity variance), it is considered as a higher IID setting.

Our novel data partitioning framework encompasses both of these mechanisms. The design of label distribution skew in this study has been taken from similar experiments in the FL field \cite{li2021federated} and has been integrated within our novel data partitioning methodology, as shown in Algorithm~\ref{alg:klabelallocation}, \ref{alg:distribution_samples} and \ref{alg:dataqskew}.

\begin{algorithm*}
\caption{k Label allocation to all devices \cite{li2021federated}}\label{alg:klabelallocation}
\begin{algorithmic}
\Function{k\_label\_allocation}{$k, K, D$} \Comment{$k$: number of allocated labels}
    \State \Comment{$K$: total number of labels}
    \State \Comment{$D$: number of devices}
    \For{device d=0,1,..,D}
        \State Randomly select k labels from K such that there is no repetition
        \State Store \{d: k\} in the device:label allocation dictionary D\_K
    \EndFor
    \State \Return{D\_K}
\EndFunction
\end{algorithmic}
\end{algorithm*}

\begin{algorithm*}
\caption{Distribution of samples with replacement}\label{alg:distribution_samples}
\begin{algorithmic}
\Function{Sample\_distribution\_with\_replacement}{$D\_K, s$} 
    \State \Comment{$D\_K$: Device:label allocation dictionary }
    \State \Comment{$s$: maximum samples per device}
    \For{label k=0,1,..,K}
        \State d\_k = count of d which contain k from D\_K
        \State k\_n = total number of k samples / d\_k
        \State Store {k:k\_n} in the label sample count dictionary K\_N
    \EndFor
    \For{device d=0,1,..,D}
        \For{allocated label k=0,1,..,k\_n}
            \State Assign k\_n samples of label k to device d from K\_N
        \EndFor
        \State randomly shuffle \& split device d samples into d\_train/d\_val/d\_test (80:10:10)
        \State Store {d: d\_train} into train set dictionary D\_TRAIN
        \State Store {d: d\_val} into validation set dictionary D\_VAL
        \State Store {d: d\_test} into test set dictionary D\_TEST
    \EndFor
    \For{train device d\_train=0,1,..,D\_TRAIN}
        \If{count(d\_train) $>$ s}
            \State randomly shuffle samples and remove $\Delta s$ from d\_train
        \ElsIf{count(d\_train) $<$ s}
            \State randomly add $\Delta s$ samples from the dataset to d\_train based on D\_K 
        \EndIf
    \EndFor
    \State \Return{D\_TRAIN, D\_VAL, D\_TEST}
\EndFunction
\end{algorithmic}
\end{algorithm*}

\begin{algorithm*}
\caption{Data Quantity Skew}\label{alg:dataqskew}
\begin{algorithmic}
\Function{Data\_quantity\_skew}{$D\_TRAIN, var$} 
    \State \Comment{$D\_TRAIN$: train set dictionary }
    \State \Comment{$var$: data quantity variance threshold}
    \For{train device d\_train=0,1,..,D\_TRAIN}
        \State d\_var = randomly sample one integer within (var, 1)
        \State shuffle samples and remove (1 - d\_var) sample proportion from d\_train
    \EndFor
    \State \Return{D\_TRAIN}
\EndFunction
\end{algorithmic}
\end{algorithm*}

\subsubsection{Label Distribution Skew}

\begin{figure}
     \centering
     \begin{subfigure}[b]{0.48\textwidth}
         \centering
         \includegraphics[width=\textwidth]{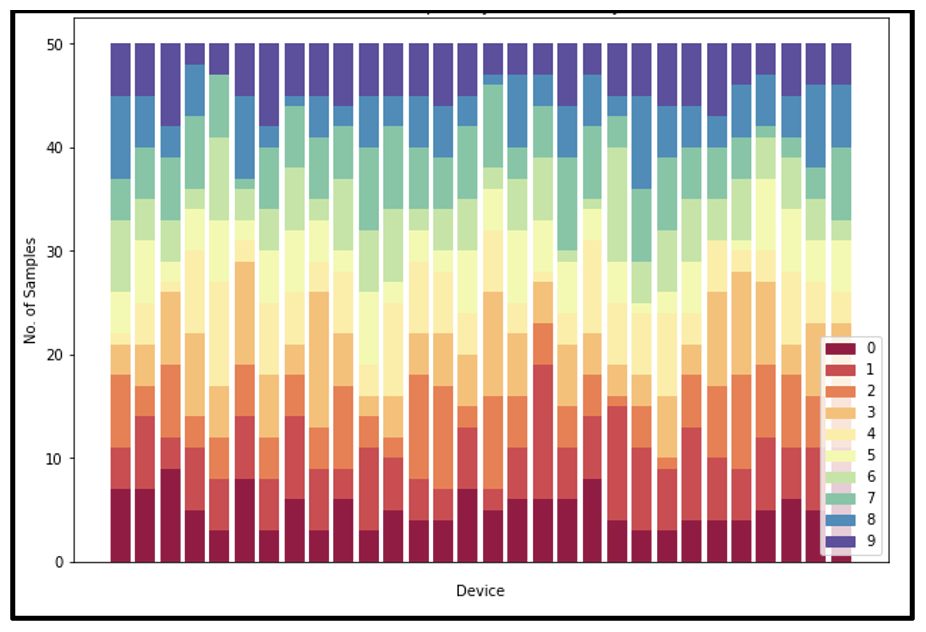}
         \caption{Label Distribution k=10}
         \label{fig:label_dist_k10}
     \end{subfigure}
     \hfill
     \begin{subfigure}[b]{0.48\textwidth}
         \centering
         \includegraphics[width=\textwidth]{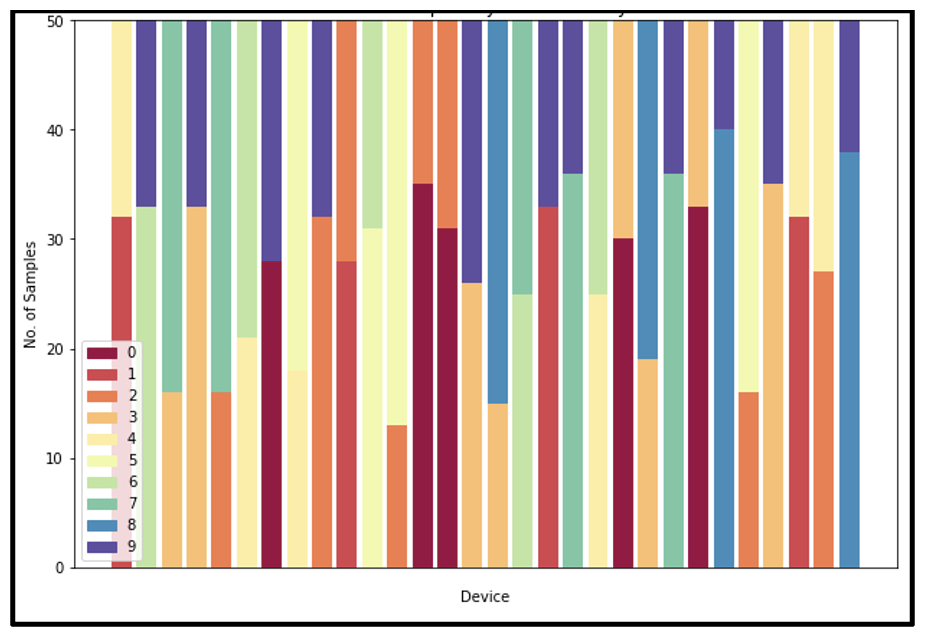}
         \caption{Label Distribution k=2}
         \label{fig:label_dist_k2}
     \end{subfigure}
        \caption{Examples of high-IID and low-IID label distribution allocations}
        \label{fig:label_dist_k}
\end{figure}

The label distribution methodology is shown in detail in pseudocode format in Algorithm~\ref{alg:klabelallocation} and \ref{alg:distribution_samples}. To fulfil the skewness of the label distribution, we first take a parameter for the number of distinct labels (k)  that each device will receive. We randomly allocate k unique labels to each device. At this stage, no data have been assigned. Once each device is allocated its respective k labels, the function will then evenly partition out all available samples from a given dataset to each device, so that there are no sample repeats across all devices. The justification for this is to ensure a fair distribution of labels from the original dataset amongst all the devices in the network. Finally, using the methodology described in Section~\ref{section:methods_methods}, we take a random 80/10/10 train/validation/test split before continuing with any further processing to produce data quantity skewness. Figure~\ref{fig:label_dist_k} shows different outputs of the label distribution process. In a high IID setting (Figure~\ref{fig:label_dist_k10}), each device contains as many distinct labels as there are all available labels in the dataset, while in the highly skewed low IID setting (Figure~\ref{fig:label_dist_k2}), each device is assigned only 2 labels. Each client, shown as a column in the figure, is only able to train and test on the samples it has been provided.

\subsubsection{Data Quantity Skew}

The next step in the data partitioning methodology controls how much data each device receives and is shown as pseudocode in Algorithm~\ref{alg:dataqskew}. This part of the data partitioning functions takes in a parameter for the number of samples per device to determine the threshold of training samples which each device will be allocated. Devices that are over this threshold will have samples randomly removed, whilst devices that are under the threshold will undergo sampling with replacement (from samples with labels that the device has been allocated) from the global dataset. Since data partitioning comes before training, there is no violation of FL privacy principles because there will be no sharing of samples once training has started \cite{li2020federated}.

\begin{figure}
     \centering
     \begin{subfigure}[b]{0.48\textwidth}
         \centering
         \includegraphics[width=\textwidth]{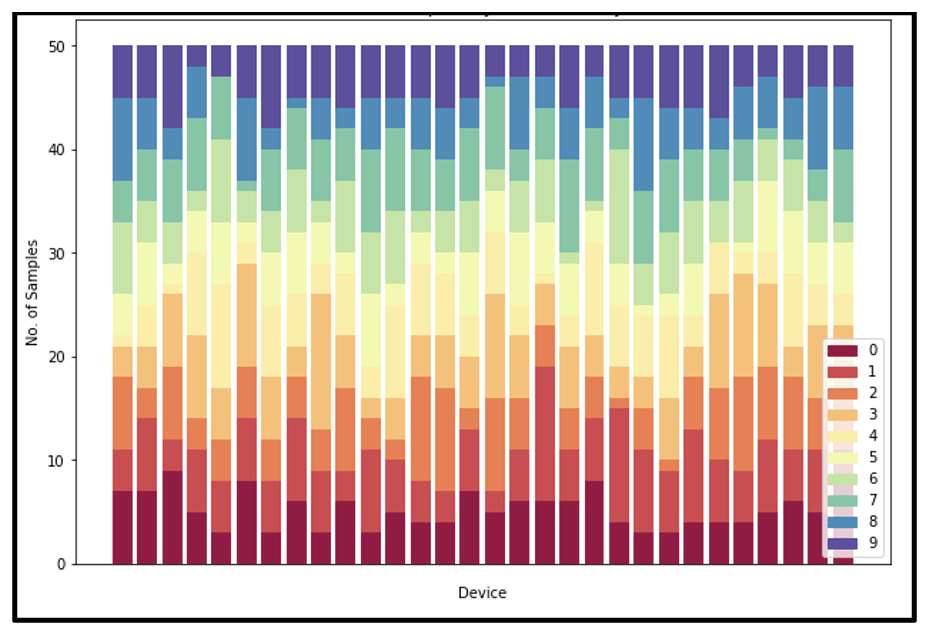}
         \caption{Data Quantity Variance = 0\%}
         \label{fig:data_variance_0}
     \end{subfigure}
     \hfill
     \begin{subfigure}[b]{0.48\textwidth}
         \centering
         \includegraphics[width=\textwidth]{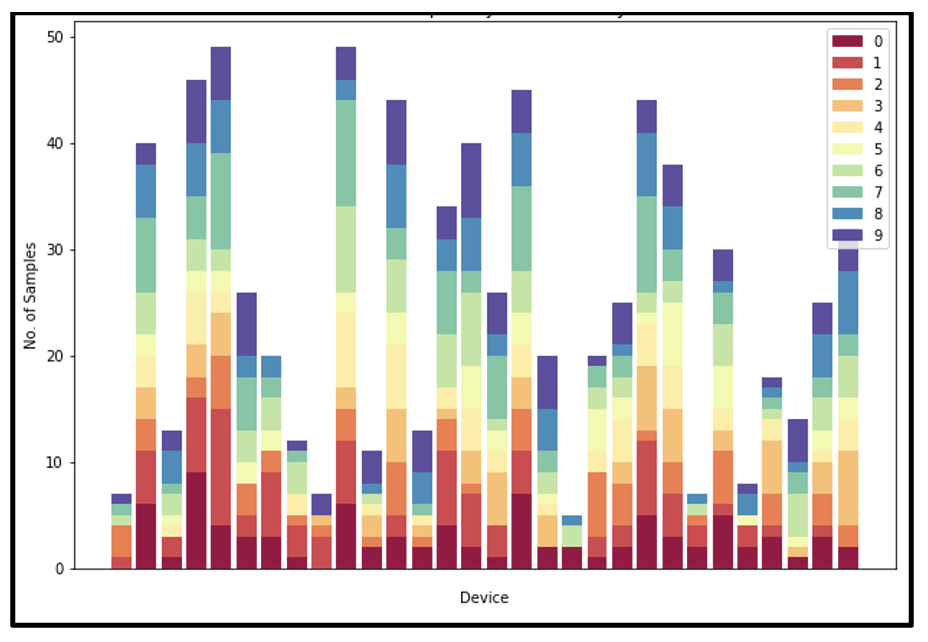}
         \caption{Data Quantity Variance = 90\%}
         \label{fig:data_variance_90}
     \end{subfigure}
        \caption{Examples of low and high variance data quantity distribution allocations}
        \label{fig:data_variance}
\end{figure}

Additionally, this function will also take in a parameter for the data quantity variance to determine the threshold of samples to be removed from each device, hence inducing a data quantity skew. Each device is randomly allocated a proportion of samples to be kept and a portion that will be removed via a threshold parameter. For example, if a device has randomly been assigned a threshold value of 0.7, then 30\% of the samples on the device will be removed at random, and this constitutes a setting of 30\% data quantity variance. Figure~\ref{fig:data_variance} shows different outputs of the data quantity skewness while fixing the label distribution. In the low-variance setting (Figure~\ref{fig:data_variance_0}), each device contains all available labels, whilst in the highly skewed setting (Figure~\ref{fig:data_variance_90}), some devices may have up to 90\% of samples removed.

\subsubsection{Quantifying Overall Data Skew}

After the allocation of samples of multiple classes and varying amounts, we next quantify how much variation there is among the entire system using the Earth-Mover distance metric (EMD) \cite{zhao2018federated}. EMD is the size of the distance in probability variation of two distributions of classes, and can be used in an FL setting when the underlying distribution of classes on each local device varies. As such, the difference in EMD between the distributions of classes on all the local devices in the system quantifies the degree of IID of the system. An array of literatures \cite{zhao2018federated,liu2020secure} have used EMD to measure the disparity between local devices and therefore quantify the degree of data skewness. Being a distance-based metric, higher values of EMD represent higher degrees of data skewness \cite{zhao2018federated}.

In the context of this research, each distribution is defined as the list of labels assigned to each local client after the data partition process. To calculate the EMD value of our federated network of clients, we compare the distributed labels of all combinations of local clients after data partitioning. Then we take the average of all pairs of EMD scores to calculate the average EMD for that data partitioning condition. Table~\ref{table:emd_analysis_mnist} shows an example of the effect of how EMD changes with label skewness, that is, the number of labels assigned to the local client, while the quantity of data skew was kept constant.

\begin{table}
\caption{EMD analysis on MNIST dataset}
\centering
\label{table:emd_analysis_mnist}
\begin{tabular}{@{}ccccc@{}}
\toprule
Number of & Number of & Data quantity & Max Number & \multirow{2}{*}{EMD} \\ 
labels & clients & threshold & of samples & \\
\midrule
1 & 30 & 0.7 & 400 & 3.2 \\
2 & 30 & 0.7 & 400 & 2.8 \\
3 & 30 & 0.7 & 400 & 2 \\
4 & 30 & 0.7 & 400 & 1.9 \\
5 & 30 & 0.7 & 400 & 1.6 \\
6 & 30 & 0.7 & 400 & 1.2 \\
7 & 30 & 0.7 & 400 & 0.9 \\
8 & 30 & 0.7 & 400 & 0.8 \\
9 & 30 & 0.7 & 400 & 0.5 \\
10 & 30 & 0.7 & 400 & 0.2 \\
\bottomrule
\end{tabular}
\end{table}

Intuitively, we can see that as the number of classes assigned to local clients increases while all other parameters are kept identical, the EMD decreases as the degree of label skewness has been reduced. For a 10-class label dataset such as MNIST, the highest EMD score fluctuates to a value of approximately 4, and this occurs in a single pair of clients when only one class has been assigned to each local device. This aligns with our expectation, as more class labels means more chances for similarities between local clients and therefore a reduced EMD value. However, due to the effect of random sampling and quantity skewness, in our study, EMD does not ever reach 0, although this is the metric's lower bound. In our experimentation, there will always be some differences in terms of labels between local clients. 

\begin{figure}
\centering
\includegraphics[width=0.7\textwidth]{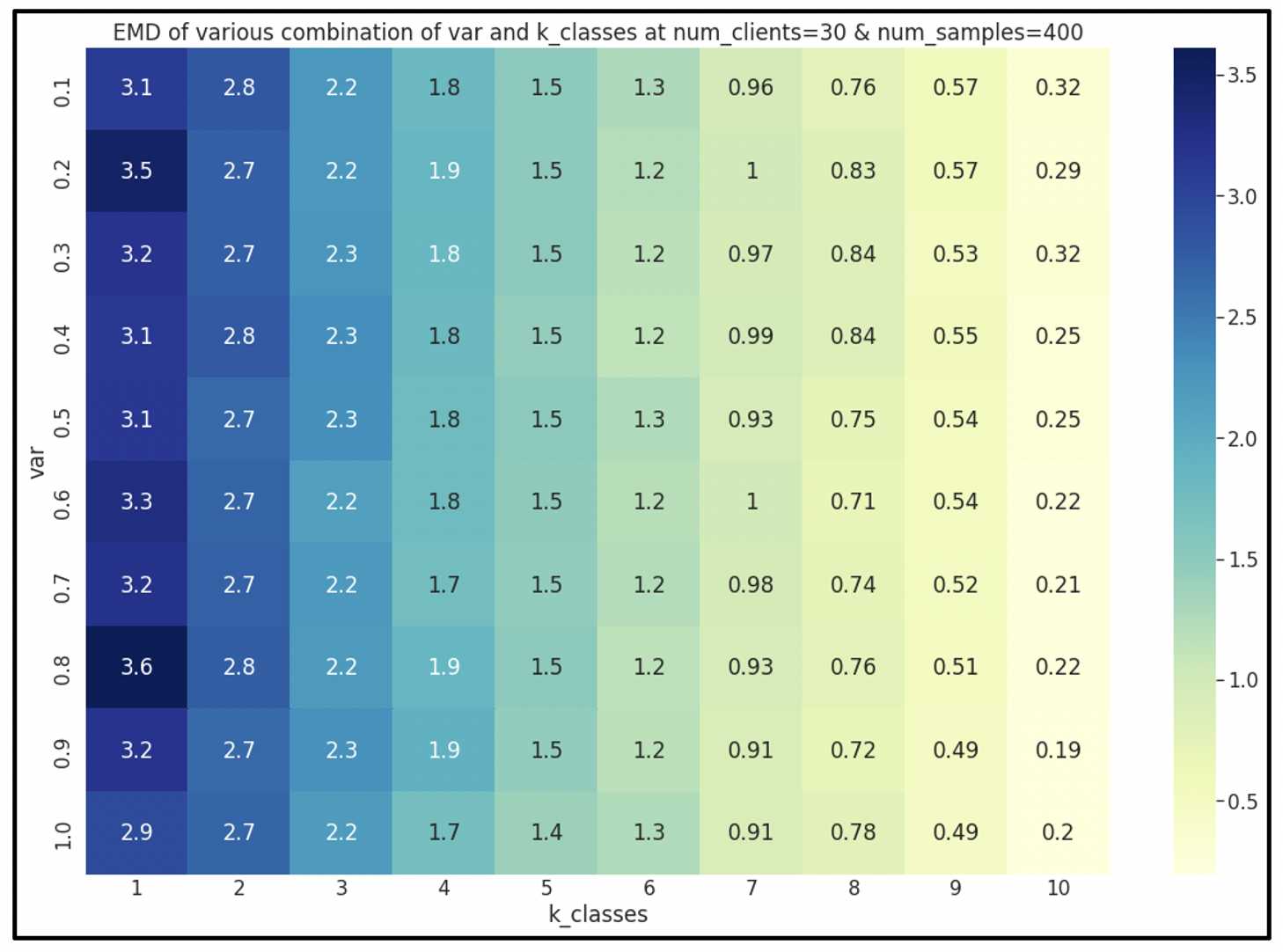}
\caption{EMD analysis using MNIST}
\label{fig:emd_analysis_mnist}
\end{figure}

Additionally, we attempted to quantify the impact of how data quantity skewness impacted EMD along with label distribution skewness. When both are taken into account, it is observed that the label distribution is the most influential parameter in changing the statistical heterogeneity of the data, as shown in Figure~\ref{fig:emd_analysis_mnist}. The heat map allows investigation of the interaction between the variance in sample quantity (var) and the number of labels assigned to local clients (\textit{k\_classes}) in the output EMD score. It is also observed that data quantity variance has some influence in changing the level of statistical heterogeneity; however, the influence is inconsistent and not as clear when compared to the impact of label distribution.

\begin{figure}
\centering
\includegraphics[width=0.7\textwidth]{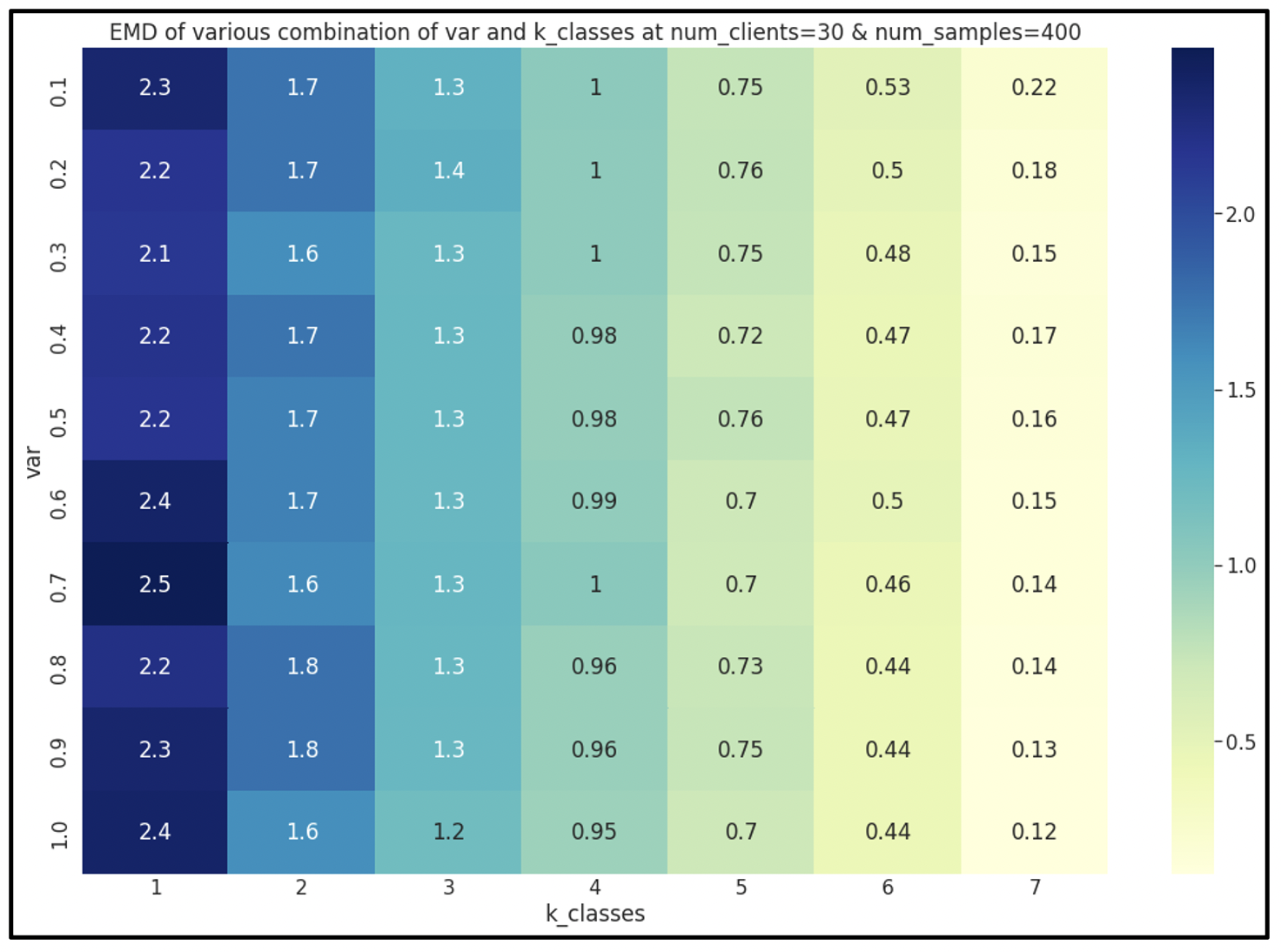}
\caption{EMD analysis using Healthcare Obesity dataset}
\label{fig:emd_analysis_healthcare}
\end{figure}

Similar results were held for the CIFAR-10 dataset, which, like MNIST, contains 10 class labels. Unlike MNIST and CIFAR-10, the Healthcare dataset contains only 7 classes. Using the same analysis approach to investigate the interaction between the distribution of labels, the quantity skewness and the EMD values on the Healthcare dataset is shown in Figure~\ref{fig:emd_analysis_healthcare}, it is evident that the maximum EMD score is lower than when using 10 class labels. With a maximum score of 2.3, this impacts how we characterise the degree of data skewness in the federated system when using datasets with different numbers of distinct class labels.

Our literature review suggests that the performance of a federated model deteriorates as the data become more statistically heterogeneous \cite{zhao2018federated,wang2020optimizing}. Importantly, we can now quantify different statistically heterogeneous settings. To validate the level of data skewness using EMD in our experiment, we ran some preliminary experiments using the four different aggregators, in order to better understand the impact of statistically heterogeneous settings on aggregator performance. Ideally, if EMD could quantify degrees of IID, and if different degrees of IID were to have an impact on aggregator performance (measured by F1 scores), we would observe decreasing F1 scores as EMD increased. If such categories are to exist, we might expect to see distinct "elbow points" where F1 scores drop substantially. Additionally, if the variance in data quantity had an impact, we would see an increase in the variance in the F1 scores with an increase in the variance in data quantity.

\begin{figure}
     \centering
     \begin{subfigure}[b]{0.48\textwidth}
         \centering
         \includegraphics[width=\textwidth]{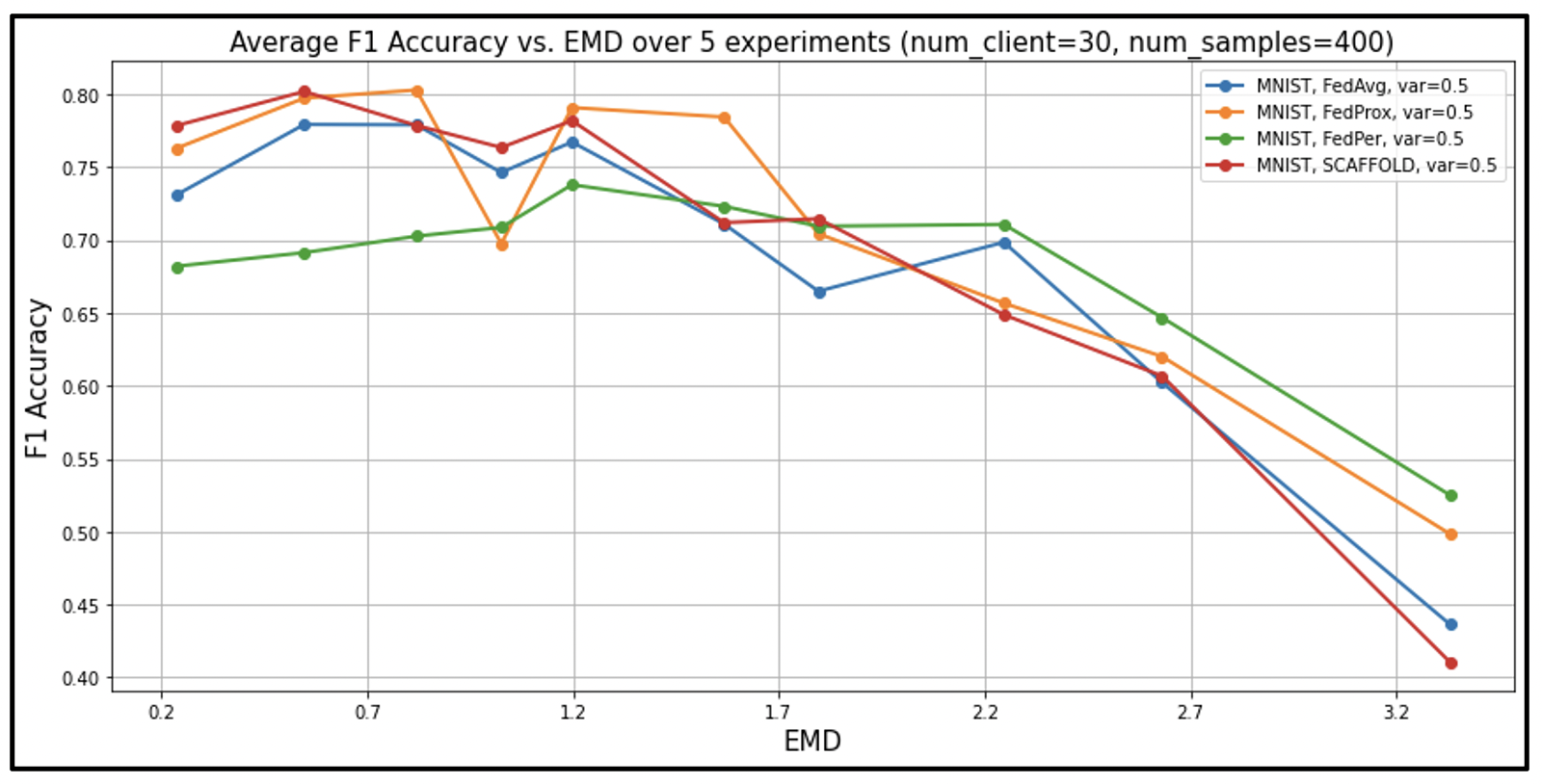}
         \caption{Average F1 accuracy vs. EMD over 5 trials, with 50\% data quantity variance}
         \label{fig:f1_accuracy_vs_emd_mnist_50}
     \end{subfigure}
     \begin{subfigure}[b]{0.48\textwidth}
         \centering
         \includegraphics[width=\textwidth]{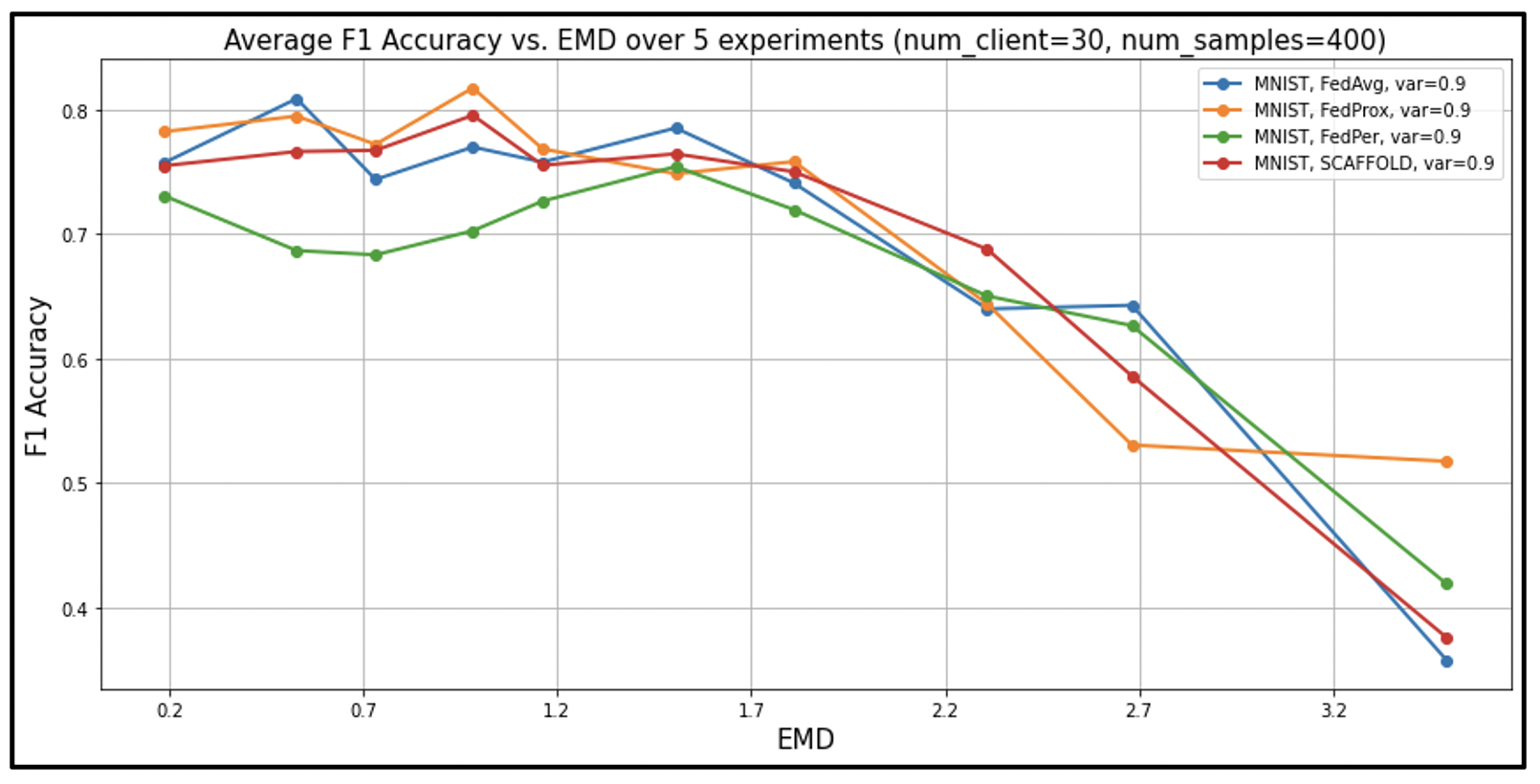}
         \caption{Average F1 accuracy vs. EMD over 5 trials, with 10\% data quantity variance}
         \label{fig:f1_accuracy_vs_emd_mnist_10}
     \end{subfigure}
        \caption{F1 accuracy vs. EMD using MNIST across two levels of data quantity variance}
        \label{fig:f1_accuracy_vs_emd_mnist}
\end{figure}

As seen in Figure~\ref{fig:f1_accuracy_vs_emd_mnist}, when there is more variance in the quantity of data, in the lower ranges of the EMD values in each figure, there is more variance in the F1 scores. With increased variance in the quantity of data, for example, variance in the quantity of data 50\%, as shown in Figure~\ref{fig:f1_accuracy_vs_emd_mnist_50}, increasing EMD values are associated with increased volatility in F1 scores. This is possibly due to fluctuation in the number of training samples available in local models. Therefore, increasing the degree of statistical heterogeneity along with increased data quantity variance is associated with increased variance in F1 scores.

Even with increased volatility in F1 scores, there appears to be a pattern that characterises distinct IID settings by their EMD values and F1 scores. Following the elbow points, when the EMD is less than 1.2, the F1 scores are high in value and stable. As the EMD becomes larger than 1.2, the F1 scores start to experience a large drop, representing the first 'elbow point'. Following that, another large drop is observed when the EMD becomes larger than 2.2.  By these three ranges of EMD values, we classify each as low, moderate and high IID respectively, as shown in Table~\ref{table:emd_thresholds_values_k_labels}.

\begin{table}
\caption{EMD Thresholds and Values of k Labels for each IID Setting}
\centering
\label{table:emd_thresholds_values_k_labels}
\begin{tabular}{@{}ccccc@{}}
\toprule
\textbf{Levels of} & \textbf{Number of } & \multirow{2}{*}{\textbf{EMD range}} & \textbf{Label} & \textbf{k Labels}  \\ 
\textbf{IID} & \textbf{labels} & & \textbf{distance} & \textbf{Value} \\
\midrule
\multirow{2}{*}{Low} & 10 & \textgreater{}2.2 & 1,   2 & 2 \\ 
 & 7 & \textgreater{}1.7 & 1,   2 & 2 \\ \midrule
\multirow{2}{*}{Medium} & 10 & Between 1.2 and 2.2 & 3-6 & 5 \\ 
 & 7 & Between   1 and 1.7 & 3,   4 & 4 \\ \midrule
\multirow{2}{*}{High} & 10 & \textless{}1.2 & 7-10 & 10 \\
 & 7 & \textless{}1 & 5-7 & 7 \\ \bottomrule
\end{tabular}
\end{table}

\begin{figure}
\centering
\includegraphics[width=0.48\textwidth]{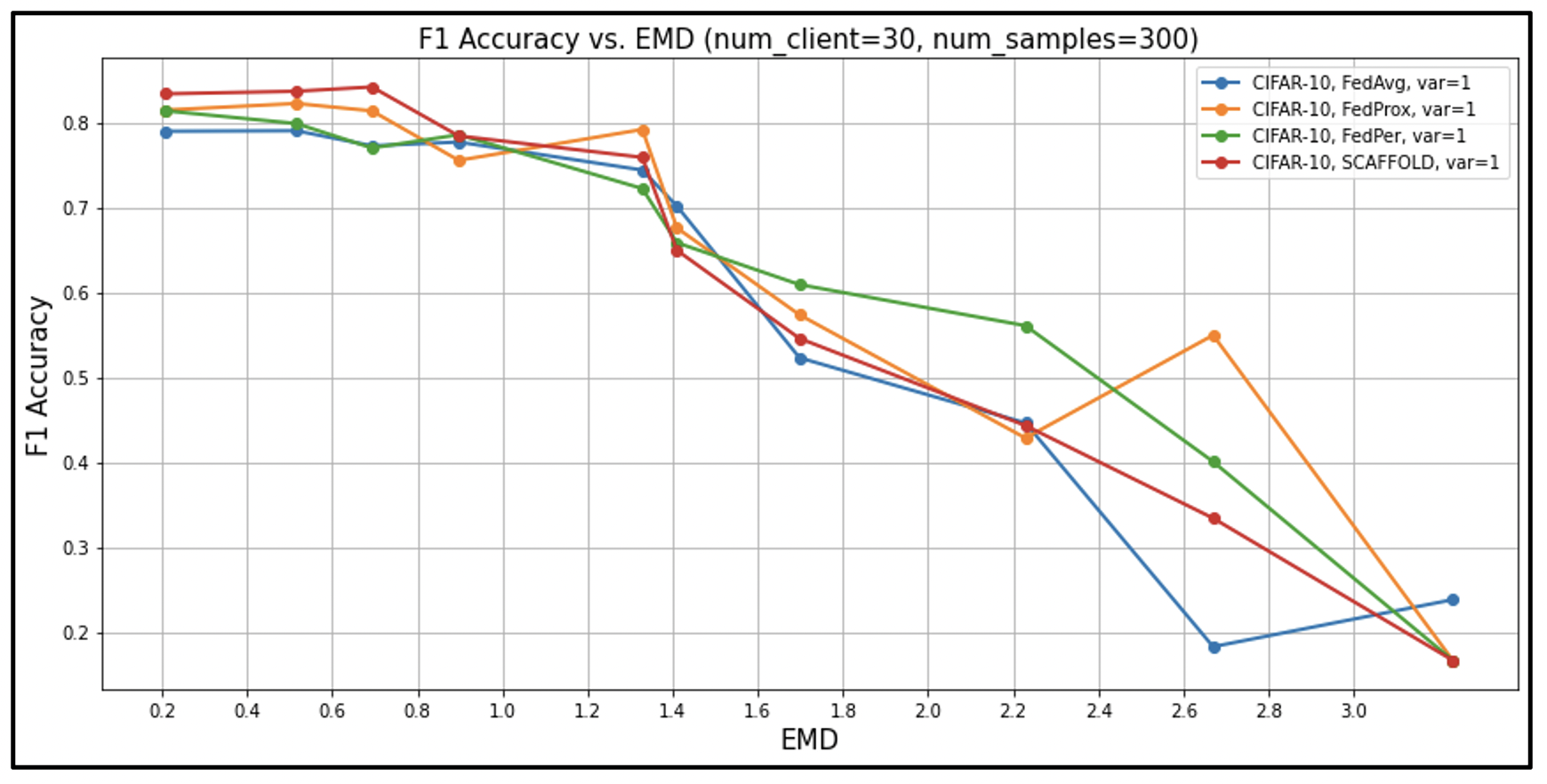}
\caption{F1 accuracy vs. EMD using CIFAR-10}
\label{fig:f1_accuracy_vs_emd_cifar10}
\end{figure}

\begin{figure}
\centering
\includegraphics[width=0.48\textwidth]{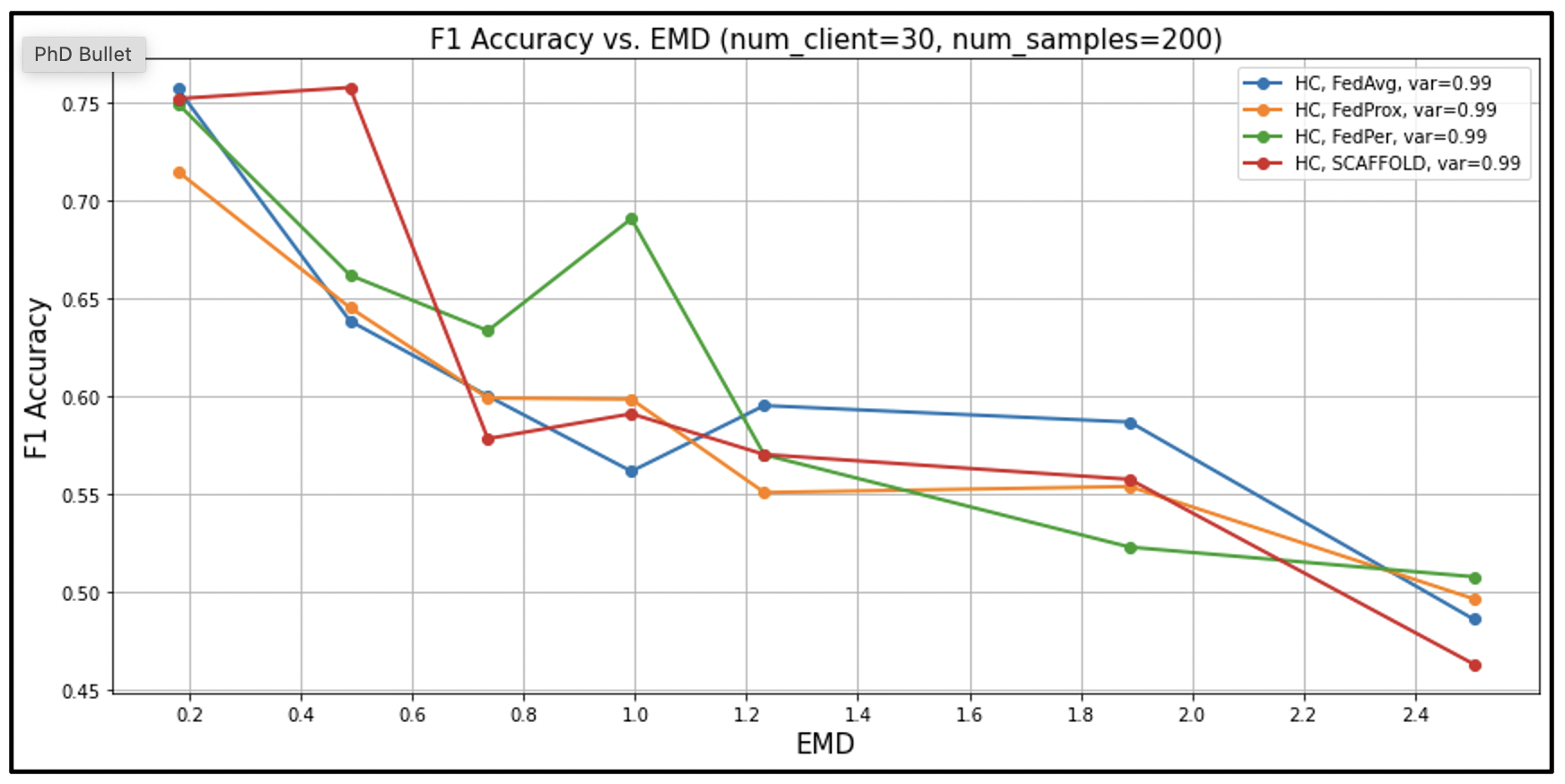}
\caption{F1 accuracy vs. EMD using Health Care Obesity dataset}
\label{fig:f1_accuracy_vs_emd_healthcare}
\end{figure}

We repeated the analysis in MNIST above using the CIFAR-10 and Healthcare datasets, as shown in Figures~\ref{fig:f1_accuracy_vs_emd_cifar10} and~\ref{fig:f1_accuracy_vs_emd_healthcare}. Although this time with 0\% data quantity variance, the same elbow points appear to hold for CIFAR-10. This is likely due to the fact that CIFAR-10 has 10 class labels, the same as MNIST.

Similarly, while using the elbow point approach for the Healthcare dataset, despite the presence of some outlier results in Figure~\ref{fig:f1_accuracy_vs_emd_healthcare} (e.g., FedProx at EMD = 1), we observed that the F1 accuracy experiences a drop for the EMD values between 0 and 1 and a large drop after EMD = 1.7. Therefore, we establish the Table~\ref{table:emd_thresholds_values_k_labels} to conclude the relationship between the IID levels and the class distribution.

Taking into account the time and resource constraints when running several FL experiments, and given that similar F1 scores were observed within each category of IID values, we selected a single value of k labels to run as a proxy for the entire range of labels that constitute an IID category. These values are shown in the column "Selected Value of k Labels" in Table~\ref{table:emd_thresholds_values_k_labels}. It is worth noting that previous research suggests that assigning only one class to local devices in a low IID setting is not suitable for running federated learning experiments, as FedAvg and FedProx will become very unstable, or even completely unable to learn, as was the case with SCAFFOLD \cite{li2021federated}.

\subsection{Experiment Setup}\label{section:methods_methods}

In each experiment, we will compare the performance of 4 aggregators: FedAvg, FedProx, FedPer, and SCAFFOLD. In evaluating results, FedAvg, as the original FL aggregator, will be considered as the baseline aggregator with which the other aggregators are compared. FedAvg is chosen as the baseline aggregator as it is a straightforward algorithm. The aggregation involves  performing training locally on each client, and averaging the model updates at each client. Due to its simplicity, FedAvg is often used as the starting point and baseline for comparison with other aggregation algorithms in federated learning. Before introducing different types of data skew, first we take each dataset in full, and split it into training and testing sets. 

From the training set, we further split it into training and validation sets. Each client is assigned data that are partitioned from these sets. We will use the testing set to evaluate the performance of the aggregators after training and aggregation has been performed in each communication round. One communication round involves training on selected devices, model aggregation, sending the aggregated model to all devices, and testing. To further simulate the impact of real-world FL, we randomly select a subset of local clients, which is an idea that was inspired by research by Cho, JJ, Wang, J, and Joshi, G \cite{cho2020client}. In their paper, they performed a convergence analysis of federated optimisation and measured the impact of client selection strategies and selection bias on the speed of convergence. Based on their study, they found a client selection technique to achieve faster error convergence by prioritising clients with higher local loss. This inspired our experimental design, as we adopted a similar approach to client selection.

In all experiments, we consider the below parameters as relevant to developing and testing the FL framework:
\begin{itemize}
    \item The number of active devices randomly selected from the total number of available devices.
    \item Local training epochs used by the devices.
    \item Batch size used by the devices.
\end{itemize}

In addition, we decided on the values for the number of communication rounds used in total by the system for each dataset, the learning rates, the number of samples each device receives, and the values for the parameters specific to each aggregator. We discuss this further in Section~\ref{section:methods_parameters}.  To isolate the impact of each parameter on the FL setting, we performed ablation studies, keeping all other parameters consistent and varying only that parameter. We conducted these experiments in three datasets: two image recognition datasets (MNIST and CIFAR-10) that are well-researched in the literature, and one external tabular dataset (Health Care Obesity dataset) which poses a novel problem.

Each dataset has an associated neural network model. Specifically, for MNIST, we have used an enhanced version of the LeNet-5 Convolutional Neural Network model \cite{lecun1989handwritten}. The model has two convolutional layers, followed by a pooling layer twice with 32 filters and 64 filters, respectively, and three fully connected layers with a SoftMax unit in the end with 10 classes. This is similar to the model used by McMahan et al. in MNIST \cite{mcmahan2017communication}.
For the more complex image recognition tasks found in the CIFAR-10 dataset, we have used MobileNetV2, which is a lightweight network that uses residual bottleneck layers to limit the number of model parameters \cite{sandler2018mobilenetv2}. MobileNetV2, as its name suggests, was designed for use on handheld mobile devices and is therefore relevant for FL experimentation. This decision was driven by resource constraints, since models with fewer parameters are likely to be faster to train, although at the expense of benchmark results seen in the literature.

To further validate our FL models, we sourced an external healthcare dataset (hereafter referred to as Healthcare dataset), authored by Palechor and de la Hoz Manotas \cite{palechor2019dataset}. This dataset is relevant to the FL data privacy objective, since data in industry may be confined to many siloes \cite{xu2021federated}, such as hospitals. This dataset aims to predict the estimated level of obesity based on variables such as gender, age, height, weight, and family history of being overweight. Obesity levels are divided into seven different classes ranging from 1-7. We have used a simple, multi-layer perceptron (MLP) model that makes use of batch normalisation, dropout and ReLU activation for classification. The reasons of using MLP model is that the model is well-suited for handling categorical variables and can effectively classify data into multiple classes \cite{brouwer2002feed}. The dropout and ReLU activation can be used to prevent overfitting and introduce non-linearity into the model, respectively \cite{dahl2013improving}.

\begin{figure}
\centering
\includegraphics[width=0.4\textwidth]{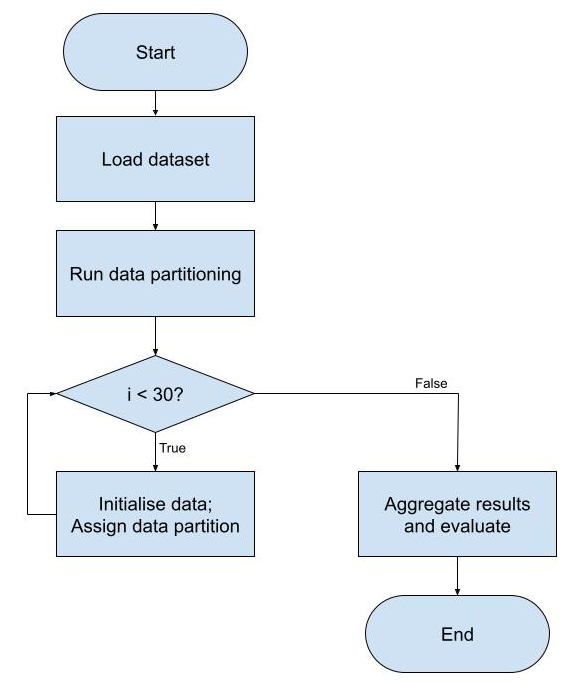}
\caption{Diagram of the overall experimental process}
\label{fig:overall_experiment_diagram}
\end{figure}

After implementing the four FL aggregators and collecting the data, we simulated 30 devices with their respective local models and a central client that aggregates the global model. To begin the FL learning process, we applied our novel data partitioning method to the three datasets in order to vary the degree of IID amongst the network. The resulting EMD value when using our partitioning function is deterministic and is based on the input values such as number of labels, number of clients, number of samples, etc. This results in the EMD value becoming repeatable, and will be the same as a result of the indices and labels of samples selected by the function, which are also deterministic in a seeded environment. Practically, this improves the robustness of each permutation of the experiment condition and allows reproducible experimentation and results.
Having set the degree of IID for the data in the network, we keep the output samples constant while varying the values of other parameters, such as local epochs and batch size, which are consequential for FL in our experiment. The overall process is described in Figure~\ref{fig:overall_experiment_diagram}.

As observed in the FL literature, results are visually displayed in terms of F1 scores by communication rounds.
To analyse results, we built functions to display results in a reproducible way, and also to record the best F1 scores over all communication rounds.
%
%
As outlined in Section~\ref{section:methods_methods}, experimentation in this study involves different combinations of FL parameter values such as the number of active devices, local training epochs, and batch size. We selected two values for each of these parameters to test: one lower and one higher. Importantly, as we are able to keep all other parameters constant, as well as the IID setting, we are then able to fairly compare the performance of aggregators by their F1 scores.

\subsection{Hyperparameters for Experiments}\label{section:methods_parameters}

To select the values of the parameters for experimentation, we were informed both by a review of the literature and by our own experimentation about what was practical, reasonable, and impactful.

\subsubsection{Number of devices}

A total of 30 devices were available for selection during the FL training process in all experiments. During preliminary experiments and investigations into data partitioning and EMD, we determined that a reasonably large number of devices was necessary, relative to the smaller number of devices that has been cited in the literature, for example, 10 \cite{arivazhagan2019federated}. The reasoning for this was that when only 10 devices are sampled and when data quantity variance is high, there may be an excessive amount of samples that are not used at all from the original dataset. On the other hand, if the total number of devices is too high and the data quantity variance is low, samples will be frequently resampled with replacement throughout the network. This could impact results as a consequence of overfitting. No consistent value for this parameter was reported in our review of the literature, with some experiments allowing up to 100 devices across the network \cite{karimireddy2019scaffold}. Taken together, we found that the practical number of devices is 30, which provides a healthy balance between sample utilisation and sample replacement.

\subsubsection{Number of active devices}

In the original FedAvg paper \cite{mcmahan2017communication},  a relatively comprehensive investigation was carried out that offered insight into reasonable values for this parameter: five levels of fractions of participating devices were tested. Other aggregators did not reliably test this parameter. Based on this, we chose the number of active devices as 20\% and 50\% of the total available devices. With 30 total devices, this meant values of 6 and 15 active devices. For the MNIST and Healthcare datasets, this was practical. Full device participation was not tested as it was reported \cite{mcmahan2017communication} that there were little or no performance gains to full participation. However, running a single experiment with CIFAR-10 would take approximately 71 minutes when using 15 active devices. Following this, to appropriately test CIFAR-10 in a practical way, we instead use values of 10\% (3) and 20\% (6) active devices to improve run times to be as short as 5 minutes.

\subsubsection{Data quantity variance}

After analysing the results of the experiment on the EMD and F1 scores, as shown in Section~\ref{section:methods_datapartition}, it was observed that the "elbow points" of the F1 scores in multiple values of data quantity variance were relatively consistent. The impact of data quantity variance appeared primarily to be the increase in the variation in F1 scores rather than the radical alteration of the IID category. Therefore, in an effort to reduce the scope of experimentation so as to alleviate time and resource constraints, we opted to conduct all of our experiments using a data quantity variance value of 0\%. After this, we conducted a small investigation of data quantity variance again to confirm the effects of EMD on F1 scores, but using the best aggregators from each combination of IID category and dataset.

\subsubsection{Number of samples per client}

We were guided by the original FedAvg paper \cite{mcmahan2017communication} for the number of samples each client would receive. It was reported that in the case of MNIST, 600 samples per client were used. In order to find a smaller number of samples to assign per client, we found that we were able to achieve similar accuracy scores using half as many samples. In the same paper, CIFAR-10 was tested with 500 samples per client. We opted to use 600 samples because our experimentation achieved similar F1 scores more slowly than what was presented in the literature. For the Healthcare dataset, unlike MNIST and CIFAR-10 with tens of thousands of samples, it contained 2100 training samples. Hence, we selected a smaller number, with 200 samples to be allocated to each device.

\subsubsection{Number of communication rounds}

For MNIST, we chose 100 communication rounds, similar to how many were used in the original FedProx paper \cite{li2020federatedB}. Although CIFAR-10 has been presented in the literature with up to 3000 communication rounds \cite{mcmahan2017communication}, Li et al. displayed F1 scores similar to what we were able to achieve with 50 communication rounds \cite{li2021federated}.  For the Healthcare dataset, 200 communication rounds were selected, as initial experiments showed that after 200 communication rounds, F1 scores tended to plateau.

\subsubsection{Learning rate}

Although research has suggested that learning rate is an important factor to consider in a federated setting \cite{hsu2019measuring}, our initial experimentation did not reveal substantial interactions, so we opted to use values cited in the literature. For MNIST and CIFAR-10, we chose a learning rate of 0.01 and 0.001 respectively. We then confirmed that these values were appropriate based on the literature reviewed \cite{mcmahan2017communication}. We were unable to find studies that have trained models using the Healthcare dataset, so it posed a novel challenge in finding a suitable learning rate. Among the values of 0.01 and 0.001, 0.001 demonstrated the best F1 scores.

\subsubsection{Number of local epochs}

\begin{figure}
\centering
\includegraphics[width=0.48\textwidth]{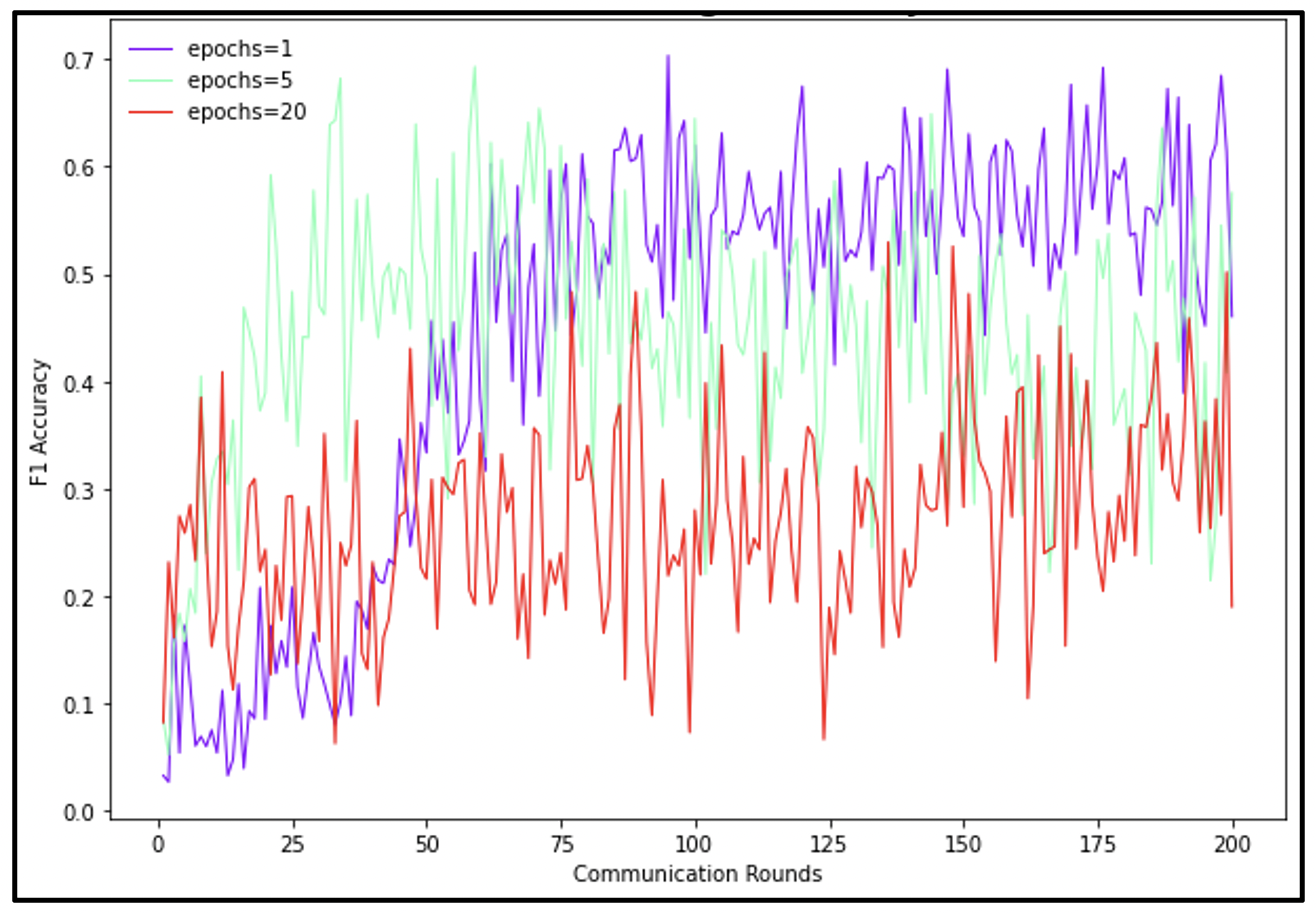}
\caption{FedAvg Healthcare Epoch Testing}
\label{fig:healthcare_epoch_testing}
\end{figure}

The selected number of local epochs used is not consistent in the literature. In the FedAvg paper \cite{mcmahan2017communication}, MNIST was tested using values for the number of local epochs as 1, 5 and 20. Given resource and time constraints, we, therefore, opted to use 1 and 5 epochs.  Furthermore, for the Healthcare dataset, initial experimentation demonstrated that performance decreased when epochs increased beyond 5 (Figure~\ref{fig:healthcare_epoch_testing}). Therefore, 20 epochs were deemed unnecessary for further experimentation. 

For CIFAR-10 experiments, similar experiments in the literature have used 4 epochs \cite{arivazhagan2019federated}, which is comparatively low. However, with 50 communication rounds and 4 local epochs, the experiment duration for a single setting takes approximately 60 minutes, while using 2 epochs takes approximately 23 minutes. Therefore, we opt to use the values of 1 and 2 for CIFAR-10.

\subsubsection{Batch size}

The literature is also not consistent in selecting the values for batch sizes. For MNIST, batch sizes of 10 and 50 have been previously tested \cite{mcmahan2017communication}, and therefore were selected. For consistency, these batch sizes were also applied to the Healthcare dataset. For CIFAR-10, a batch size of 128 is often cited as an appropriate batch size \cite{arivazhagan2019federated,mcmahan2017communication}. To create the lower parameter value, this was halved to become 64.

\subsubsection{Aggregator Parameters} 

In addition to FedAvg, each of the other aggregators has a unique parameter that influences the mechanism of the aggregator. In FedProx, the term mu determines the influence of the proximal term on the regularisation of the objective. It is suggested that any value greater than zero is effective, and we take the recommendation to set the parameter at a value of 0.001 \cite{li2020federatedB}. For FedPer, the number of personalised layers was shown to have no particular correlation with performance, but at least one layer was effective \cite{arivazhagan2019federated}. As such, we used one layer of personalisation. Finally, for SCAFFOLD, the global learning rate is a term that scales the degree of influence that updated local models have on the global model \cite{karimireddy2019scaffold}. No reason is offered for tuning this value or if doing so would even be desirable. We kept the global learning rate at a value of 1.0.

Lastly, a decision was made not to experiment with FedProx on CIFAR-10 due to runtime constraints. In the worst case, a single experiment with FedProx on CIFAR-10 would take 4.5 hours. This decision was justified by the results of Li et al. \cite{li2021federated}, who also noted that the large computational overhead of FedProx is due to the modification of the objective function that requires the norm to be taken of all layers of weights, resulting in a slow and complex operation.


\section{Results}

Having performed the methodologies described in the previous section, we outline the results of our experiments. Each section breaks down the experimental results by the parameters that were varied: batch size, active devices, and local epochs. Within each section, we describe the effects of each of the parameters tested by IID setting: highly IID, moderately IID, and low-IID.

\subsection{Active Devices}

As shown in Table~\ref{table:results_active_devices_high}, in the high IID setting, the best F1 scores were observed in all three datasets with fewer active devices than more. When using fewer devices, there was only a small increase in the best F1 scores compared to the use of more devices on the MNIST and CIFAR-10 datasets. However, in the case of the Healthcare dataset, which was a tabular dataset, there was a clear disadvantage to selecting more active devices.

\begin{table}
\caption{Aggregator results by number of active devices across all datasets in the high IID setting}
\centering
\label{table:results_active_devices_high}
\begin{tabular}{cccccc}
\toprule
 & \textbf{Active} & \multicolumn{4}{c}{\textbf{Aggregator}} \\ \cline{3-6}
\textbf{Dataset} & \textbf{Devices} & \textbf{FedAvg} & \textbf{FedProx} & \textbf{FedPer} & \textbf{Scaffold} \\
\midrule
MNIST & 6 & 98.4 & \textbf{98.6} & \textbf{98.2} & \textbf{98.5} \\
MNIST & 15 & 98.4 & 98.2 & 98.2 & 98.5 \\
CIFAR-10 & 3 & 50.9 & - & 50.3 & \textbf{52.9} \\
CIFAR-10 & 6 & 49.1 & - & 47.6 & 52.2 \\
Healthcare & 6 & 82.5 & 79.7 & \textbf{85.7} & 83.5 \\
Healthcare & 15 & 74.3 & 73 & 73 & 74.1 \\
\bottomrule
\end{tabular}
\end{table}

Across the three datasets, there was no clear trend toward the best aggregator in the high IID setting. The best F1 score was recorded by FedProx (on MNIST), FedPer (on Healthcare) and SCAFFOLD (on CIFAR-10). FedAvg however ranked second best for each dataset.

\begin{table}
\caption{Aggregator results by number of active devices across all datasets in the moderate IID setting}
\centering
\label{table:results_active_devices_moderate}
\begin{tabular}{cccccc}
\toprule
 & \textbf{Active} & \multicolumn{4}{c}{\textbf{Aggregator}} \\ \cline{3-6}
\textbf{Dataset} & \textbf{Devices} & \textbf{FedAvg} & \textbf{FedProx} & \textbf{FedPer} & \textbf{Scaffold} \\
\midrule
MNIST & 6 & \textbf{99.1} & 98.8 & 98.5 & 99 \\
MNIST & 15 & 98.6 & 98.7 & 98.4 & 98.9 \\
CIFAR-10 & 3 & \textbf{52} & - & 50.2 & 49.4 \\
CIFAR-10 & 6 & 47.6 & - & 44.7 & 47.3 \\
Healthcare & 6 & 80 & 81.3 & \textbf{87.8} & 83 \\
Healthcare & 15 & 73.4 & 67.9 & 66.9 & 70.9 \\
\bottomrule
\end{tabular}
\end{table}

In the moderately IID setting, and again in all three datasets, fewer active devices gave the best F1 results (Table~\ref{table:results_active_devices_moderate}). Compared to the high IID setting, there was a greater difference in the best F1 scores between the larger and smaller number of active devices. In addition to having two of the three best F1 scores in each dataset, FedAvg was also the best scoring aggregator in most experiments in the moderately IID setting, regardless of whether more or fewer devices were used. FedAvg's scores were also more similar between the two conditions when compared to other aggregators. By comparison, SCAFFOLD similarly showed lower variance between the best F1 scores in the lower and higher device conditions, and FedPer achieved the best results in the Healthcare dataset.

\begin{table}
\caption{Aggregator results by number of active devices across all datasets in the low IID setting}
\centering
\label{table:results_active_devices_low}
\begin{tabular}{cccccc}
\toprule
 & \textbf{Active} & \multicolumn{4}{c}{\textbf{Aggregator}} \\ \cline{3-6}
\textbf{Dataset} & \textbf{Devices} & \textbf{FedAvg} & \textbf{FedProx} & \textbf{FedPer} & \textbf{Scaffold} \\
\midrule
MNIST & 6 & \textbf{99.3} & 99.2 & 99 & 99.2 \\
MNIST & 15 & 99 & 98.8 & 98.4 & 98.7 \\
CIFAR-10 & 3 & \textbf{70.4} & - & 64.4 & 64.3 \\
CIFAR-10 & 6 & 39.7 & - & 43.6 & 48.4 \\
Healthcare & 6 & \textbf{78.7} & 76 & 69.7 & 78.5 \\
Healthcare & 15 & 59.3 & 55.8 & 59.4 & 58.5 \\
\bottomrule
\end{tabular}
\end{table}

In the low IID setting, fewer active devices again demonstrated the best scores in all three datasets (Table~\ref{table:results_active_devices_low}). FedAvg was the best performing aggregator, achieving the best score on all three datasets. However, FedAvg also had the largest performance gap between the two device conditions. Compared to high IID and moderately IID settings, in low IID settings, there were greater gaps in F1 scores between the number of active devices for each aggregator and dataset. 

\begin{figure}
\centering
\includegraphics[width=0.48\textwidth]{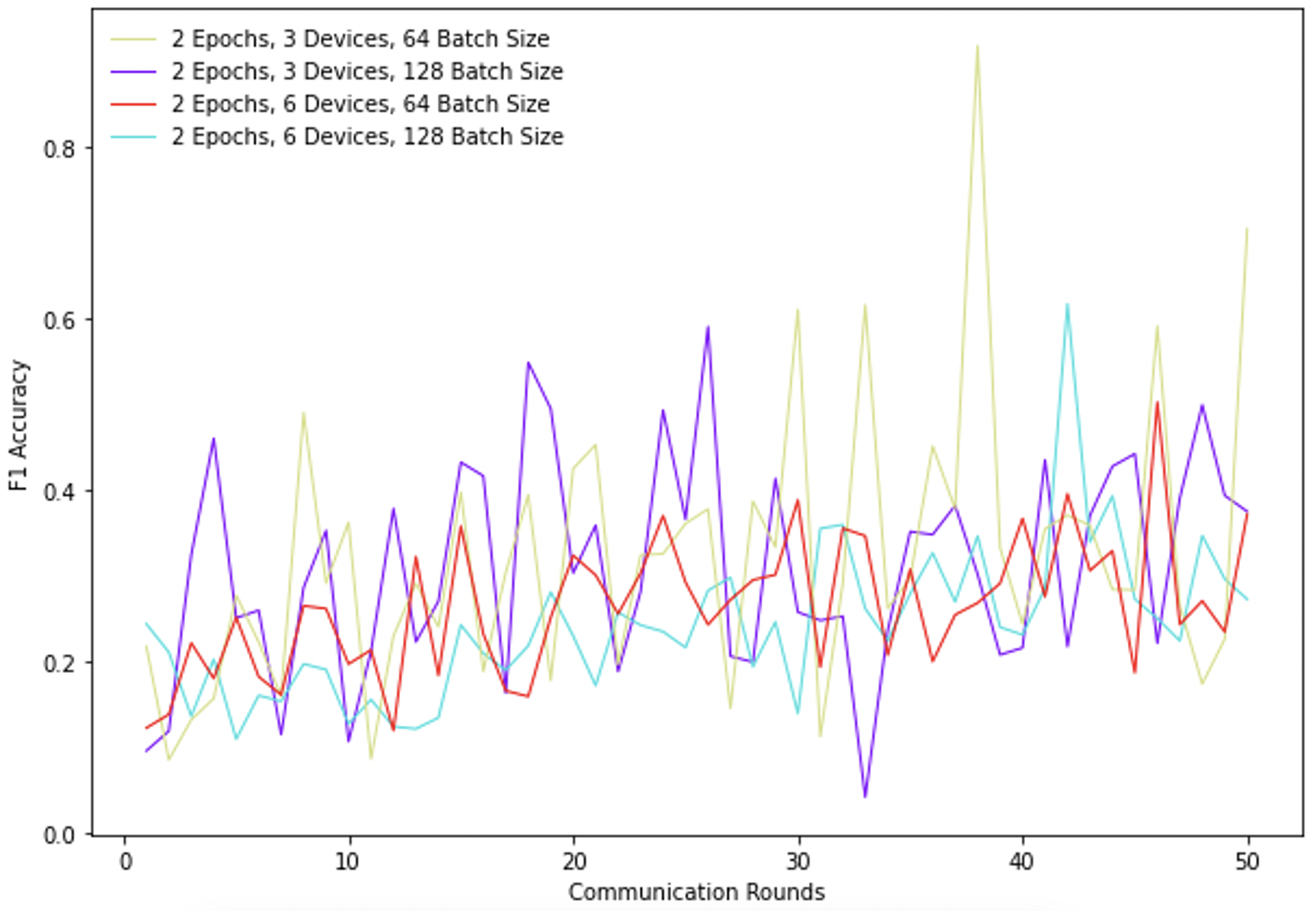}
\caption{Evidence of outliers and volatility in FedAvg experiments on CIFAR-10, low IID setting using both batch size 64 and 128}
\label{fig:outliers_evidence_and_volatility}
\end{figure}

With an unusually high score for CIFAR-10 in the low IID setting, FedAvg was able to outperform the best results in CIFAR-10 in the moderately IID and high IID settings. Upon inspecting the learning curves (Figure~\ref{fig:outliers_evidence_and_volatility}), this result can be interpreted as the consequence of an outlier score. The low IID setting also demonstrates volatile learning curves, where there is a high variance between F1 scores in each consecutive communication round.

\begin{figure}
\centering
\includegraphics[width=0.95\textwidth]{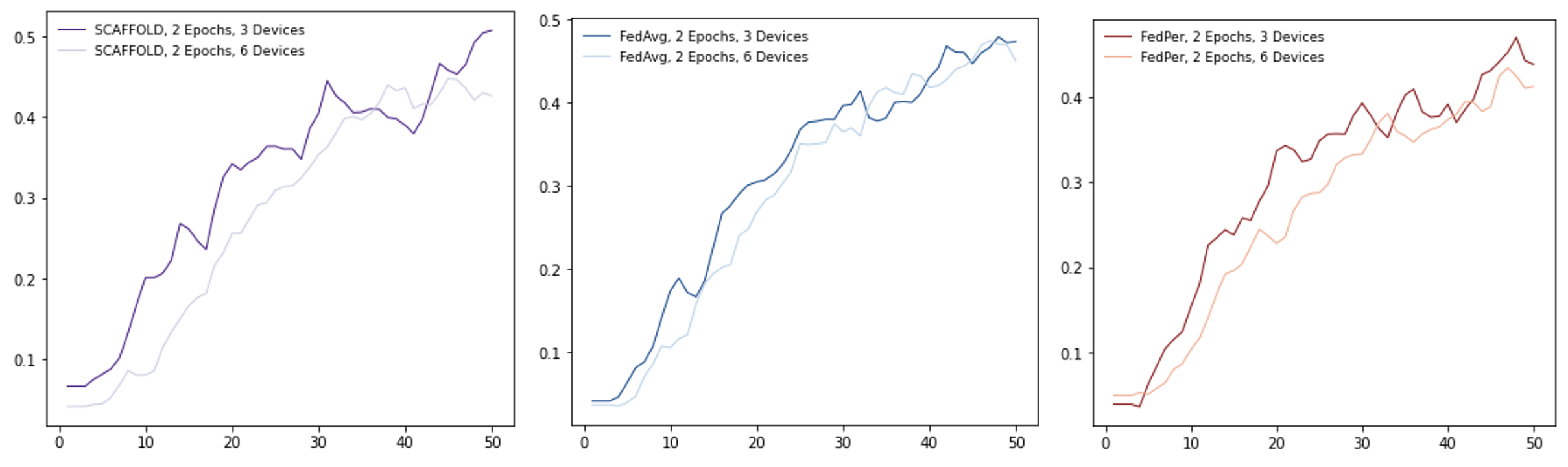}
\caption{Results from CIFAR-10 in a highly IID setting, with fewer devices (darker lines) showing better F1 scores being achieved before the same score was achieved by more active devices}
\label{fig:results_cifar10_high_iid}
\end{figure}

To find greater clarity, outliers greater than three times the standard deviation of the learning curve were removed. Volatility was also smoothed using a moving average (window size = 3), and backfilling removed outliers. After this process, it became clearer that FedAvg was prone to outliers in the low IID setting and that SCAFFOLD was the best performing aggregator in CIFAR-10 in the low IID setting, as shown in Table~\ref{table:results_active_devices_low_outliers}. 
Lastly, it was generally observed that fewer active devices resulted in higher F1 scores being achieved earlier than with fewer communication rounds, as seen in Figure~\ref{fig:results_cifar10_high_iid}.

\begin{table}
\caption{Aggregator results by number of active devices across all datasets in the low IID setting, after removing outliers and smoothing volatility on CIFAR-10}
\centering
\label{table:results_active_devices_low_outliers}
\begin{tabular}{ccccc}
\toprule
& & \multicolumn{3}{c}{\textbf{Aggregator}} \\
\cline{3-5}
\textbf{Dataset} & \textbf{Active Devices} & \textbf{FedAvg} & \textbf{FedPer} & \textbf{SCAFFOLD} \\
\midrule
CIFAR-10 & 3 & 48.3 & 43.4 & \textbf{51.7} \\
CIFAR-10 & 6 & 36.1 & 35.7 & 39 \\
\bottomrule
\end{tabular}
\end{table}

\subsection{Epochs}

The FL training results when varying the number of local training epochs in a high IID setting are shown in Table~\ref{table:results_n_epochs_high}. It can be seen that the general trend in the collected results suggests that the overall accuracy score improved with more training epochs. This was found that in most cases, when comparing the aggregator with the aggregator, the accuracy usually improved with a greater number of epochs. For example, in experiments with MNIST and CIFAR-10 as datasets, we can see that there was a consistent improvement in accuracy scores across all aggregators. FedPer and SCAFFOLD recorded an improvement of approximately 10\% in the case of CIFAR-10.

However, an interesting observation was made in experiments done for the Healthcare dataset. The accuracy of FedAvg and FedProx was better in the lower epoch condition. FedAvg recorded an accuracy drop of 4\% with the highest epoch value, while the number of epochs had no significant impact on how FedProx worked with the dataset. FedPer and SCAFFOLD gave better accuracy with more epochs, and that was in line with what was observed for the MNIST and CIFAR-10 datasets. SCAFFOLD and FedPer gave consistently better results with more epochs in a high IID setting for all 3 datasets.

\begin{table}
\caption{Aggregator results by number of epochs across all datasets in the high IID setting after removing outliers and smoothing volatility on CIFAR-10}
\centering
\label{table:results_n_epochs_high}
\begin{tabular}{cccccc}
\toprule
 &  & \multicolumn{4}{c}{\textbf{Aggregator}} \\
\cline{3-6}
\textbf{Dataset} & \textbf{Epochs} & \textbf{FedAvg} & \textbf{FedProx} & \textbf{FedPer} & \textbf{Scaffold} \\
\midrule
MNIST & 1 & 98.3 & 98.2 & 98.1 & 98.3 \\
MNIST & 5 & 98.4 & \textbf{98.6} & 98.2 & 98.5 \\
CIFAR-10 & 1 & 46.7 & - & 41.9 & 44.2 \\
CIFAR-10 & 2 & 51 & - & 50.4 & \textbf{52.9} \\
Healthcare & 1 & 82.5 & 79.7 & 82.9 & 83.5 \\
Healthcare & 5 & 78.4 & 79.4 & \textbf{85.7} & 77.1 \\
\bottomrule
\end{tabular}
\end{table}

Table~\ref{table:results_n_epochs_moderate} shows the results obtained from varying numbers of training epochs in a moderately IID environment. In contrast to what was observed for a high IID setting, in a moderate IID setting, it was observed that 2 out of 3 datasets (MNIST and Healthcare) performed better with fewer epoch values.
Experiments carried out with the Healthcare dataset showed a notable improvement in accuracy of approximately 30\% accuracy when using fewer epochs. In the case of MNIST, there was not much difference in the performance of the aggregators under different epoch conditions, but the best overall result was achieved with fewer epoch values. However, in the case of CIFAR-10, most of the experiments showed that the aggregators worked better when the highest epoch value was selected.

Across all aggregators in these experiments, FedAvg performed relatively well for all data sets and obtained two of the three best accuracy scores (MNIST and CIFAR-10). Furthermore, it was observed that while FedPer was definitely not the best performing aggregator for the image recognition datasets, it was the best performing aggregator for the Healthcare dataset (a tabular dataset) outputting highest accuracy with both high and low epochs.

\begin{table}
\caption{Aggregator results by number of epochs across all datasets in the moderate IID setting after removing outliers and smoothing volatility on CIFAR-10}
\centering
\label{table:results_n_epochs_moderate}
\begin{tabular}{cccccc}
\toprule
 &  & \multicolumn{4}{c}{\textbf{Aggregator}} \\
\cline{3-6}
\textbf{Dataset} & \textbf{Epoch} & \textbf{FedAvg} & \textbf{FedProx} & \textbf{FedPer} & \textbf{Scaffold} \\
\midrule
MNIST & 1 & \textbf{99.1} & 98.8 & 98.3 & 98.6 \\
MNIST & 5 & 98.8 & 98.8 & 98.5 & 99 \\
CIFAR-10 & 1 & 48.6 & - & 47.7 & 49.4 \\
CIFAR-10 & 2 & \textbf{52} & - & 50.2 & 48.3 \\
Healthcare & 1 & 80 & 81.3 & \textbf{87.8} & 83 \\
Healthcare & 5 & 53.3 & 50.8 & 56.7 & 49 \\
\bottomrule
\end{tabular}
\end{table}

\begin{figure}
\centering
\includegraphics[width=0.55\textwidth]{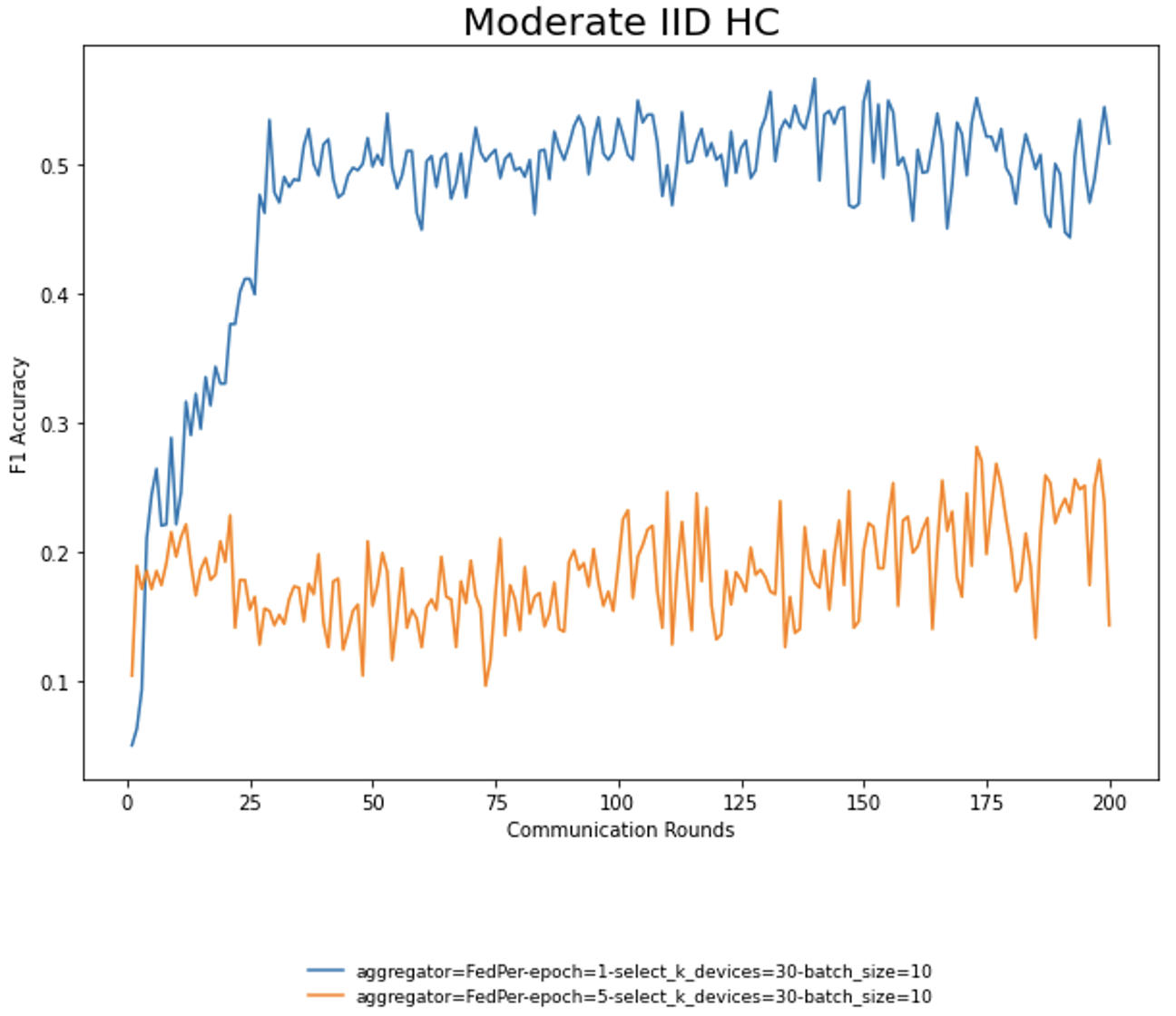}
\caption{Comparison of model performance with fewer (blue line) and greater (orange line) epoch values in moderate IID setting}
\label{fig:model_performance_moderate_iid}
\end{figure}

Figure~\ref{fig:model_performance_moderate_iid} shows the learning curve of the best observed aggregator (FedPer) for Healthcare dataset under moderate IID setting. It can clearly be seen that throughout all the communications rounds the aggregator performed significantly better with fewer epochs (blue line) as opposed to more epochs (orange line).

The experimental results conducted in low IID settings are shown in Table~\ref{table:results_n_epochs_low}. These experiments highlight that in a low IID data distribution when using MNIST and Healthcare,  the aggregators performed better with fewer epochs. In particular, experiments with the Healthcare dataset showed a significant improvement in accuracy (that is, 21. 6\% in the case of FedProx and 16. 5\% in the case of FedAvg) when the lower epoch value was selected. However, experiments using the CIFAR-10 data set did not show similar improvements with fewer epochs, and FL aggregators were found to perform best when training with more epochs. 

Among the four FL aggregators used in these experiments, FedAvg performed best in 5 out of 6 experiments. SCAFFOLD also demonstrated good results in the low IID setting and obtained results that were close to FedAvg's accuracy for the MNIST and Healthcare datasets. Interestingly, FedPer did not perform well in the low IID experiments, and achieved accuracy scores that were approximately 10\% lower than the best performing aggregator for the CIFAR-10 and Healthcare datasets.

\begin{table}
\caption{Aggregator results by number of epochs across all datasets in low IID setting after removing outliers and smoothing volatility on CIFAR-10}
\centering
\label{table:results_n_epochs_low}
\begin{tabular}{cccccc}
\toprule
 &  & \multicolumn{4}{c}{\textbf{Aggregator}} \\
\cline{3-6}
\textbf{Dataset}& \textbf{Epoch} & \textbf{FedAvg} & \textbf{FedProx} & \textbf{FedPer} & \textbf{Scaffold} \\
\midrule
MNIST & 1 & \textbf{99.3} & 99.2 & 98.7 & 99.1 \\
MNIST & 5 & 99.2 & 99.2 & 99 & 99.2 \\
CIFAR-10 & 1 & 61.5 & - & 57.2 & 57.5 \\
CIFAR-10 & 2 & \textbf{70.4} & - & 64.4 & 64.3 \\
Healthcare & 1 & \textbf{78.7} & 76 & 69.7 & 78.5 \\
Healthcare & 5 & 62.2 & 54.4 & 62.1 & 69.1 \\
\bottomrule
\end{tabular}
\end{table}

\begin{figure}
\centering
\includegraphics[width=0.55\textwidth]{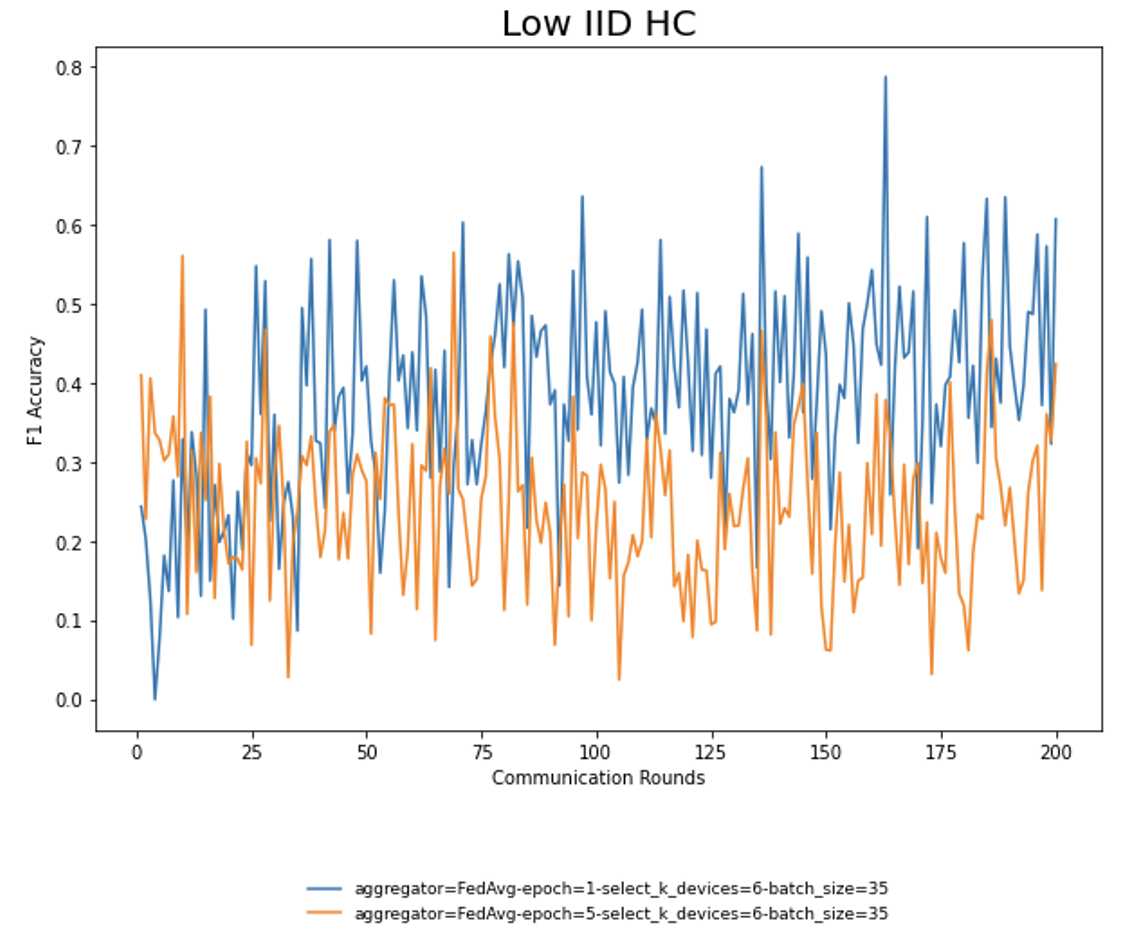}
\caption{Comparison of model performance with fewer (blue line) and greater (orange line) epoch values in low IID setting}
\label{fig:model_performance_low_iid}
\end{figure}

Figure~\ref{fig:model_performance_low_iid} shows the learning curve of the best observed aggregator (FedAvg) for Healthcare dataset under the low IID setting. Even though volatility in learning was observed, overall the aggregator performed much better with fewer epochs (blue line) throughout the communication rounds of the experiment.

\subsection{Batch Size}

In the highly IID setting, best results were observed typically with larger batch sizes, as seen in Table~\ref{table:batch_size_high}. However, the best score for MNIST was achieved with a smaller batch size using FedProx. The next best score in the larger batch size setting came from SCAFFOLD and was only slightly less than this though (98.6\% vs 98.5\%).

\begin{figure}
\centering
\includegraphics[width=0.95\textwidth]{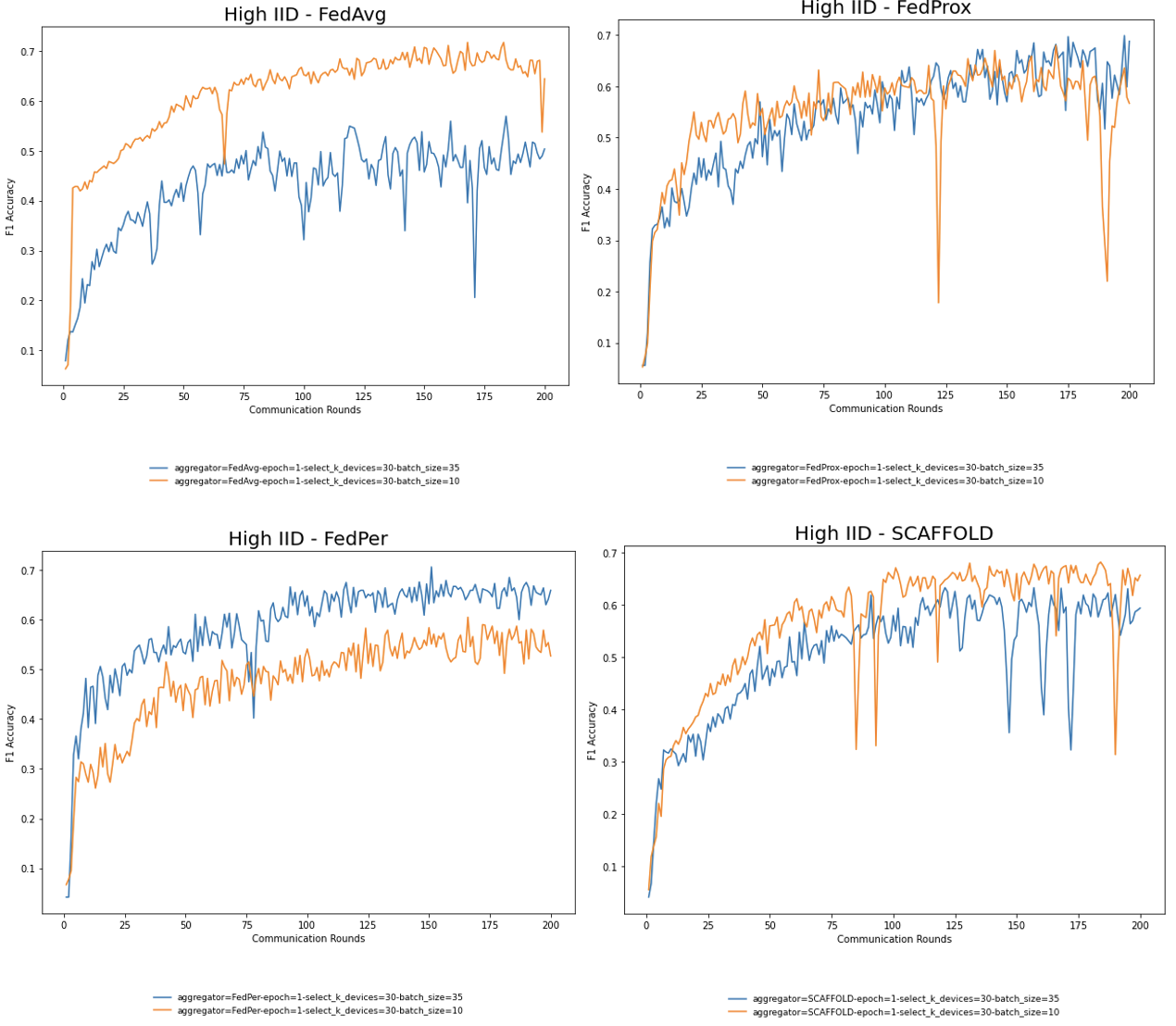}
\caption{Comparison of batch size by aggregator Healthcare dataset in high IID settings}
\label{fig:batch_size_aggregator}
\end{figure}

When observing the performance over communication rounds for the Healthcare dataset, the most stable aggregators did not offer the best accuracy. When keeping all other experimental parameters the same, we see from Figure~\ref{fig:batch_size_aggregator} there is a preference towards the smaller batch size of size 10 (orange line) than a larger batch size of size 35 (blue line) for most aggregators. This provides further support to the notion that a smaller batch size is preferred in a high IID setting.

\begin{table}
\caption{Aggregator results by batch size across all datasets in highly IID settings}
\centering
\label{table:batch_size_high}
\begin{tabular}{cccccc}
\toprule
 & \textbf{Batch} & \multicolumn{4}{c}{\textbf{Aggregator}} \\
\cline{3-6}
\textbf{Dataset} & \textbf{size} & \textbf{FedAvg} & \textbf{FedProx} & \textbf{FedPer} & \textbf{Scaffold} \\
\midrule
MNIST & 10 & 98.4 & 98.6 & 98.2 & 98.5 \\
MNIST & 50 & 98.4 & 98.4 & 98.2 & 98.5 \\
CIFAR-10 & 64 & 50.9 & - & 48 & 52.2 \\
CIFAR-10 & 128 & 49 & - & 50.3 & 52.9 \\
Healthcare & 10 & 77.8 & 79.7 & 83 & 80 \\
Healthcare & 35 & 82.4 & 79.4 & 85.7 & 83.5 \\
\bottomrule
\end{tabular}
\end{table}

In the moderately IID setting, larger batch sizes tended to outperform smaller batch sizes. This was not true for the Healthcare dataset though, where a smaller batch size gave the best performance.

\begin{table}
\caption{Aggregator results by batch size across all datasets in moderately IID settings}
\centering
\label{table:batch_size_moderate}
\begin{tabular}{cccccc}
\toprule
 & \textbf{Batch} & \multicolumn{4}{c}{\textbf{Aggregator}} \\
\cline{3-6}
\textbf{Dataset} & \textbf{size} & \textbf{FedAvg} & \textbf{FedProx} & \textbf{FedPer} & \textbf{Scaffold} \\
\midrule
MNIST & 10 & 98.6 & 98.8 & 98.4 & 98.9 \\
MNIST & 50 & \textbf{99.1} & 98.7 & 98.5 & 99 \\
CIFAR-10 & 64 & 48.6 & - & 50.2 & 48.3 \\
CIFAR-10 & 128 & \textbf{52} & - & 42.7 & 49.4 \\
Healthcare & 10 & 76.1 & 77.5 & \textbf{87.8} & 79.4 \\
Healthcare & 35 & 80 & 81.3 & 80.1 & 83 \\
\bottomrule
\end{tabular}
\end{table}

\begin{figure}
\centering
\includegraphics[width=0.95\textwidth]{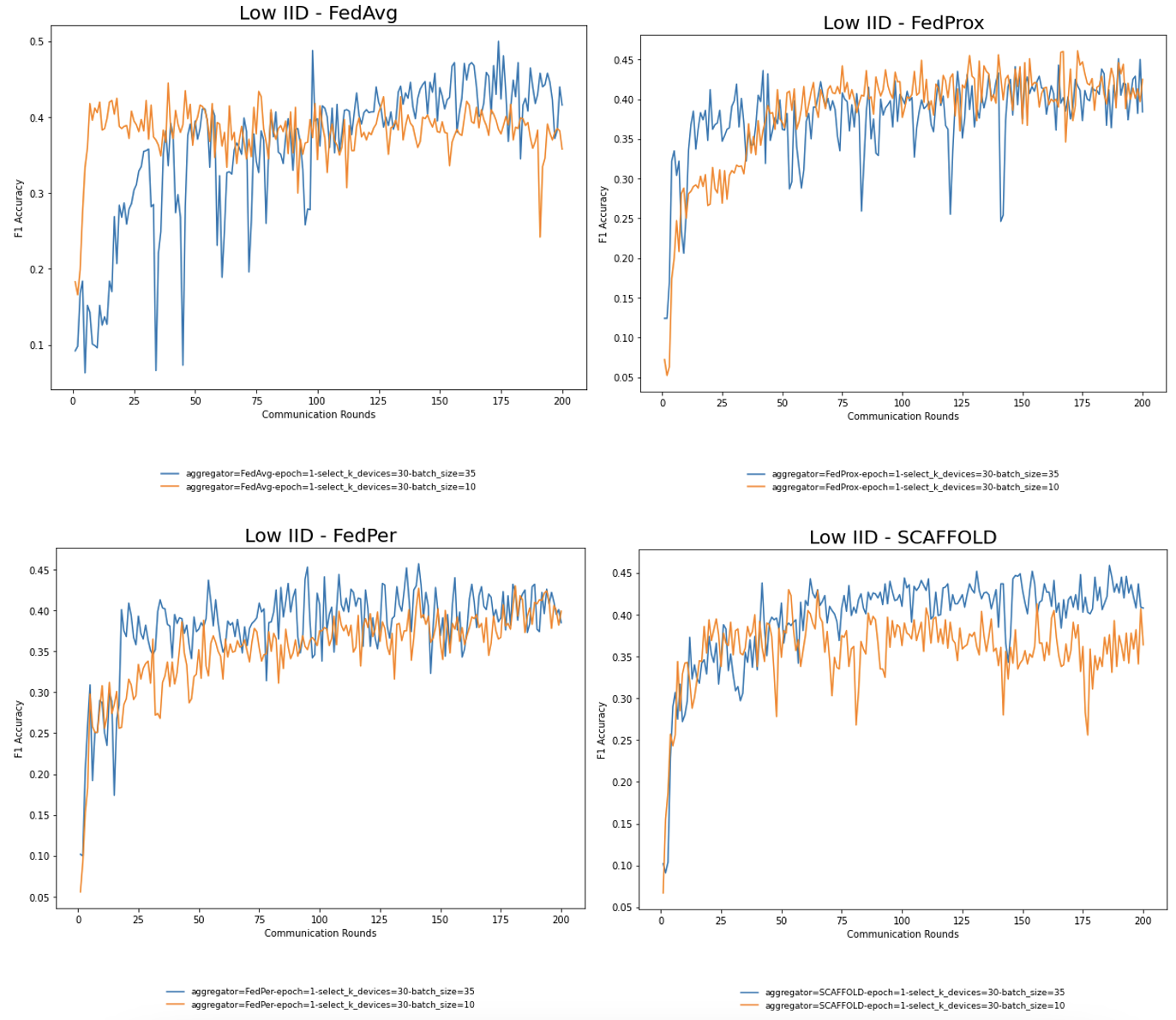}
\caption{Comparison of batch size by aggregator Healthcare dataset in low IID settings}
\label{fig:batch_size_healthcare_low}
\end{figure}

Using the MNIST and Healthcare datasets in a low IID setting, larger batch sizes performed best. However, there was no clear trend for CIFAR-10, as smaller batch sizes worked best in the low IID setting, but larger batch sizes achieved better results in the moderately IID setting. One potential explanation for this is the outliers observed in aggregator performance with CIFAR-10 in the low IID setting. The best scores for MNIST varied little between settings, although clearly a larger batch size was best. This was similar for Healthcare, as seen in Figure~\ref{fig:batch_size_healthcare_low}: when using a larger batch size of size 35 (blue line), better performance was achieved than when using a smaller batch size of size 10 (orange line).

Overall, we observed a relationship where a lower batch size is preferred in high IID settings, but a higher batch size is preferred in low IID settings. A notable deviation from this relationship was observed in FedPer and CIFAR-10,  where a larger batch size was preferred, regardless of the IID category.

\begin{table}
\caption{Aggregator results by batch size across all datasets in low IID settings}
\centering
\label{table:batch_size_low}
\begin{tabular}{cccccc}
\toprule
 & \textbf{Batch} & \multicolumn{4}{c}{\textbf{Aggregator}} \\
\cline{3-6}
\textbf{Dataset} & \textbf{size} & \textbf{FedAvg} & \textbf{FedProx} & \textbf{FedPer} & \textbf{Scaffold} \\
\midrule
MNIST & 10 & 99.1 & 99.1 & 98.6 & 99.2 \\
MNIST & 50 & \textbf{99.3} & 99.2 & 99 & 99.1 \\
CIFAR-10 & 64 & \textbf{70.4} & - & 57.2 & 64.3 \\
CIFAR-10 & 128 & 61.5 & - & 64.4 & 61.6 \\
Healthcare & 10 & 75.8 & 72.8 & 69.7 & 78.5 \\
Healthcare & 35 & \textbf{78.7} & 76 & 63 & 75.5 \\
\bottomrule
\end{tabular}
\end{table}

\subsection{Impact of Data Quantity Variance}

In addition to the results of the investigation into the variance in data quantity and EMD found in Section~\ref{section:methods_datapartition}, we provide further experimentation here on the impact of the variance in data quantity on the performance of the aggregator. As shown by Figure~\ref{fig:emd_analysis_mnist} and~\ref{fig:emd_analysis_healthcare} in Section~\ref{section:methods_datapartition}, the variance of the data quantity had little impact on the calculated EMD scores, however, the increase in the variance of the data quantity was associated with a degradation in the F1 scores. On the contrary, Li et al. \cite{li2021federated} found that there was limited impact of the data quantity skew on the aggregator performance. In this section, our aim is to further validate the effects of data quantity variance by testing the best performing aggregators (that is, in terms of best F1 scores per dataset and IID setting combination) found above against different levels of data quantity variance.

As shown in Table~\ref{table:f1score_aggregators_all}, in the case of the CIFAR-10 and MNIST datasets, a negative relationship was observed between the variance in the data quantity and the resulting F1 scores: increased variance in the data quantity was associated with decreases in the F1 score in all IID settings. However, the results in the Healthcare dataset were more inconsistent. In general, results for the Healthcare dataset became better with increased data quantity variance, although in the highly IID setting, no clear relationship was seen.

\begin{table*}
\caption{Best F1 scores from best aggregators across three levels of data quantity variance}
\centering
\label{table:f1score_aggregators_all}
\begin{tabular}{cccccc}
\toprule
&  & & \multicolumn{3}{c}{\textbf{Variance}} \\
\cline{4-6}
\textbf{Dataset} & \textbf{IID} & \textbf{Best Aggregator} & \textbf{0\%} & \textbf{50\%} & \textbf{90\%} \\
\midrule
MNIST & High & FedAvg & \textbf{98.6} & 98.4 & 98.1 \\
MNIST & Moderate & SCAFFOLD & \textbf{99.1} & 98.6 & 98.4 \\
MNIST & Low & FedPer & \textbf{98.8} & 98.3 & 98 \\
CIFAR-10 & High & SCAFFOLD & \textbf{52.9} & 41.7 & 39.2 \\
CIFAR-10 & Moderate & FedPer & \textbf{52} & 45.4 & 41 \\
CIFAR-10 & Low & SCAFFOLD & \textbf{70.4} & 64.1 & 58.2 \\
Healthcare & High & SCAFFOLD & \textbf{73.7} & 59.1 & 66.7 \\
Healthcare & Moderate & FedPer & 59.7 & 58.7 & \textbf{62.1} \\
\bottomrule
\end{tabular}
\end{table*}


\section{Discussion}

\subsection{Active Devices}

When selected at random, across all IID settings and datasets, selecting fewer devices consistently demonstrated better F1 scores than when selecting more active devices. In the low IID setting, there was a considerably greater gap between the F1 scores of the two device conditions compared to the high IID setting, where the effect was diminished. A moderate gap between the conditions was observed in the moderately IID setting, and provides further evidence to show how the effect of the number of active devices scales with each IID setting.

When varied by number of active devices, no aggregator clearly performed the best across all IID settings and datasets. However, similar to Hsu et al. \cite{hsu2019measuring}, we observed greater volatility in F1 scores in low IID settings, which may have obscured any trends in the best aggregators. Before the removal of outliers and smoothing of volatility in low IID settings, it appeared that FedAvg was capable of scoring the best results. However, after treating outliers and volatility, SCAFFOLD emerged as a strong performer in the low IID setting. SCAFFOLD appeared to have less variance between the results of the two device conditions. This supports two claims from the original SCAFFOLD paper \cite{karimireddy2019scaffold}, which concluded that the aggregator was more resilient to client sampling and that it may also be a more adequate aggregator for non-convex (i.e., low IID) problems than FedAvg.

In McMahon et al.'s original paper introducing FedAvg \cite{mcmahan2017communication}, it was reported that increasing the number of active devices corresponded to marginal increases in performance beyond a certain fraction of reporting clients (10\% of total clients). Notably, it was also shown that increasing the number of active clients did not always provide a uniform improvement. In certain scenarios, it was reported that increasing the number of active clients would reduce performance relative to smaller fractions of reporting clients. As shown in our results, we almost uniformly saw worse results when using more active devices. Additionally, McMahan et al. also reported that increasing the fraction of active devices resulted in fewer communication rounds required to reach a target level of accuracy. However, our results indicated the opposite, as shown in Figure~\ref{fig:results_cifar10_high_iid}: we observed fewer devices achieving greater F1 scores before more active devices could.

Other than the contribution from McMahan et al. \cite{mcmahan2017communication}, testing aggregators with varying numbers of active devices has been infrequently or not extensively tested. Results in literature are often reported when selecting all 10 clients from a total of 10 clients, such as with FedPer \cite{arivazhagan2019federated}, rather than with random selections of active devices. With FedProx \cite{li2020federatedB}, a related concept of stragglers is tested. A straggler is a device that is active, but has not completed the required amount of training at the time of aggregation. This concept, however, is unique to the FedProx aggregator, and reported results from varied numbers of stragglers did not indicate influence on model loss.

One explanation for the results we observed comes from Li et al. \cite{li2021federated}, who similarly recorded that better accuracy scores were obtainable by selecting fewer active devices on different aggregators. By selecting a smaller number of active devices, the aggregation of fewer models provides a clearer signal, and there is less capacity for overfitting by each individual model. Therefore, it is apparent that the selection of devices to aggregate is relevant to good global model performance. This is especially clear in lower IID settings, where the consequence of selecting a greater number of active devices can result in markedly worse global model performance.

\subsection{Epochs}

In general, it was observed that the optimal number of epochs found for a certain IID setting was influenced by the underlying data distribution. In high IID settings, the aggregators gave better accuracy with more epochs, while in moderate and low IID settings, the accuracy was better with fewer epochs. A quite substantial gap between moderate and low IID settings was observed in the scores recorded for the Healthcare dataset. This finding is also consistent across all aggregators, as shown in Figures 14 and 15. One explanation for this is that a high IID setting is more comparable to that of a centralised learning scenario, as opposed to an FL scenario. In such centralised scenarios, using more epochs typically improves the overall accuracy of a model, whereas using more epochs in a lower IID setting results in the model overfitting the data, and thus negatively impacting its accuracy. 

As shown in the results, different aggregators have different sensitivity to the number of epochs. This can be attributed to the differences in the optimisation algorithms used by each aggregator. For example, aggregators with more complex optimisation algorithms, such as SCAFFOLD, may exhibit sensitivity to the number of epochs due to the iterative nature of their learning process. On the other hand, aggregators with simpler optimization algorithms, such as FedAvg, may be less sensitive to the number of epochs as they rely more on the averaging of model updates rather than fine-tuning with each iteration. Additionally, sensitivity or insensitivity to the number of epochs can also be influenced by the inherent characteristics of the dataset. For example, datasets with a higher degree of complexity or variability may require more epochs for aggregators to converge and achieve optimal performance.

There was no single aggregator that clearly outperformed other aggregators across all IID settings and datasets. Despite this, it is worth noting that FedAvg performed better than the other aggregators in many of the lower IID settings. This observation supports the findings of Li et al. \cite{li2021federated} in which it is mentioned that the number of local epochs can have a large effect on accuracy and that the optimal value of the number of local epochs is sensitive to non-IID distributions. In lower IID settings, they reported that the accuracy decreased with higher local epoch settings. They suggested that the existing algorithms were not robust enough against large local updates. Furthermore, our findings are in line with the conclusions presented by McMahan \cite{mcmahan2017communication}, in which, under similar testing conditions, the experiment with a lower epoch value gave better results than the experiments with a higher epoch value. 

According to Wang et al. \cite{wang2020federated}, and similar to what we observed, the number of local training epochs affects the performance of aggregators such as FedAvg and FedProx as a consequence of model divergence under low IID conditions. It was also reported that only the performance of one aggregator (FedMA) improved with more local training epochs. Potentially then, certain aggregators may perform best with a number of local epochs specific for that aggregator in a given IID setting. Furthermore, the optimal number of local epochs may be variable per communication round, as suggested by Li et al. \cite{li2020federatedB}. FedProx attempts to manage this under conditions of systems heterogeneity, and though this was not specifically tested for in our experimentation, FedProx was the second best performing aggregator in lower-IID settings with fewer epochs overall. We also observed that SCAFFOLD under all IID settings always performed better with fewer epochs. This takeaway is similar to the conclusions drawn in the original SCAFFOLD paper by Karimireddy et al. \cite{karimireddy2019scaffold}.

In contrast to the findings on the MNIST and Healthcare datasets, we found that CIFAR-10 consistently had the best results when tested with more epochs rather than fewer. One explanation for this is that CIFAR-10 is a much more complex image recognition problem than MNIST, and that the difficulty of the task posed by the Healthcare dataset is comparatively less complex. Therefore, CIFAR-10 benefitted from more local epochs, and the best number of local epochs may be even greater than the values we were able to test.

Lastly, it is important to consider the goal of FL when choosing whether to increase or decrease local epochs. According to the arguments presented by Jiang et al. \cite{jiang2019improving},  the objective of FL should not be to simply produce a good global model, but rather to increase the personalised performance of local devices. They suggest that, in the case of aggregators such as FedAvg, aggregation attempts to treat decentralised data with the same objective as if it were centralised.  Consequently, improved results in lower IID settings when using fewer epochs fails to account for the goal of personalised performance. To this end, they recommend using approaches such as model-agnostic meta-learning, which may involve increasing the number of local epochs even in low IID settings.

\subsection{Batch Size}

In high IID settings, we observed the best results from smaller batch sizes. However, in low IID settings, larger batch sizes provided the best results. This is in contrast to Li et al. \cite{li2021federated}, where it was reported that batch size has no influence on aggregator performance. Our results instead somewhat support the conclusions of He et al. \cite{he2019control}, who found that in a distributed setting, batch size influences a trade-off between communication cost and convergence speed. However, they reported that large batch sizes may increase convergence speed but at the detriment of accuracy, which we did not observe in lower IID settings. Further improvements to convergence speed may come from refining the choice of optimiser, associated learning rates and values of momentum, as suggested by Felbab et al.'s work in a distributed setting \cite{felbab2019optimization}, though their results were not broken down by different levels of IID. The generalisation of convolutional neural networks has been found to be a function of the ratio between batch size and learning rate, which controls the gradient descent speed \cite{he2019control}. To this end, it has been reported that decreases in batch size can cause a comparable increase in accuracy as would be possible with a decrease in learning rates \cite{he2019control}. However, within our experiments, we assumed a constant learning rate that was specific to each data set, and therefore batch size may have played the role of a proxy for learning rate.

The preference for lower batch sizes in high IID settings can be driven by several factors. Firstly, lower batch sizes reduce communication overhead between the central server and participating devices, promoting more efficient updates. Lower batch sizes help mitigate device heterogeneity, contributing to model consistency across devices. Additionally, the use of smaller batch sizes increases model robustness by minimising the risk of overfitting to specific local patterns. In the other hand, larger batch sizes help mitigate the challenges posed by the non-uniformity and diversity of data across participating devices, facilitating more stable and representative updates to the global model. There are several implications of lower batch sizes in high IDD settings. Resource constraints on participating devices in the FL architecture are accommodated, and privacy preservation is enhanced as smaller batches reduce the potential disclosure of sensitive information during the federated learning process. Compared to in low IID settings where local data distributions vary significantly, larger batch sizes contribute to a more robust averaging of diverse gradients, enhancing the model in capturing and adapting to broader patterns in the data.

Nasirigerdeh et al. \cite{nasirigerdeh2020federated}, who offered the explanation of a generalisation gap, observed similar trends between batch size and IID settings. The generalisation gap is between local and global models, which becomes larger in low IID settings, but may be improved by using larger batch sizes on the model hosted by local clients. Additionally, our finding that smaller batch sizes achieve higher test accuracy in high IID settings has also been observed in centralised machine learning as in He et al. \cite{he2019control}, where a larger batch size instead fails to learn sharp local minima during training.

Looking more closely into the trends of aggregator performances and particular datasets, we observed that FedProx was the most robust aggregator to changes in batch size on the Healthcare dataset, as shown in Figure~\ref{fig:batch_size_healthcare_low}. This is potentially due to the choice of value for the proximal term (mu = 0.001), but is in contrast to the view of Li et al. \cite{li2021federated} where it is stated that the regularisation term of FedProx does not have a significant effect when the batch size is small. Typically, we saw similar accuracy regardless of batch size in low IID settings. FedProx algorithm incorporates a proximal term that helps prevent overfitting and promotes stability \cite{li2020federatedB}. FedProx also considers both model updates and the proximity of local models to the global model, facilitating an effective combination of information from devices with varying batch sizes. The algorithm's ability to handle data heterogeneity and encourage global consensus further contributes to its stability across devices with different training conditions. Additionally, FedProx allows for flexibility in learning rates, enabling devices to adapt to varying batch sizes during training. In summary, these characteristics make FedProx less sensitive to changes in batch size, ensuring robust performance in federated learning scenarios on the Healthcare dataset.

Regardless of the IID setting, we observed that FedPer performed the best on larger batch sizes. This suggests that the personalisation layer of the model needs sufficient batch size to train and update the base layer on each device, and perhaps aggregators such as FedProx and SCAFFOLD require even greater batch sizes to perform similarly well. Lastly, for the more complex dataset CIFAR-10, we saw that a larger batch size was preferred in all IID levels and aggregators. Interestingly, this relationship was also observed by Reyes et al. \cite{reyes2021precision}, who found that larger batch sizes were preferable for CIFAR-10, but not for MNIST, where larger batch sizes deteriorated performance. The insight drawn from this work supports the idea that there is greater complexity, or data heterogeneity, in models trained on natural images such as CIFAR-10, as opposed to the greyscale images of handwritten digits in MNIST, and this would be the case even in high IID settings. Hence, this result is sensible and expected for CIFAR-10 and highlights that there are still methods to be explored to delineate between IID levels.

\subsection{Impact of Data Quantity Variance}

We assume that F1 scores would decrease with increased data quantity variance, since the global model would be learning from less reliable data samples at each communication round. In building and validating our data partitioning method, we observed this relationship, though data quantity variance had little impact on the EMD metric when we experimented with the best aggregators, and typically this trend emerged again across all IID settings. In the case of the CIFAR-10 and MNIST datasets, increased data quantity variance was associated with worsened F1 scores. However, with the Healthcare dataset, F1 scores tended to improve with greater variance in the quantity of data. Potentially, this is influenced by the dataset's tabular nature as opposed to the more complex image recognition problems of CIFAR-10 and MNIST. Given that the results tended to improve with reduced data quantity variance and that the corresponding EMD scores did not decrease with reduced data quantity variance, we conclude that to fully capture the nature of the IID setting, the data quantity variance as a measure should be quoted in conjunction with the EMD metric.

\subsection{Summary}

\begin{figure*}
\centering
\includegraphics[width=0.98\textwidth]{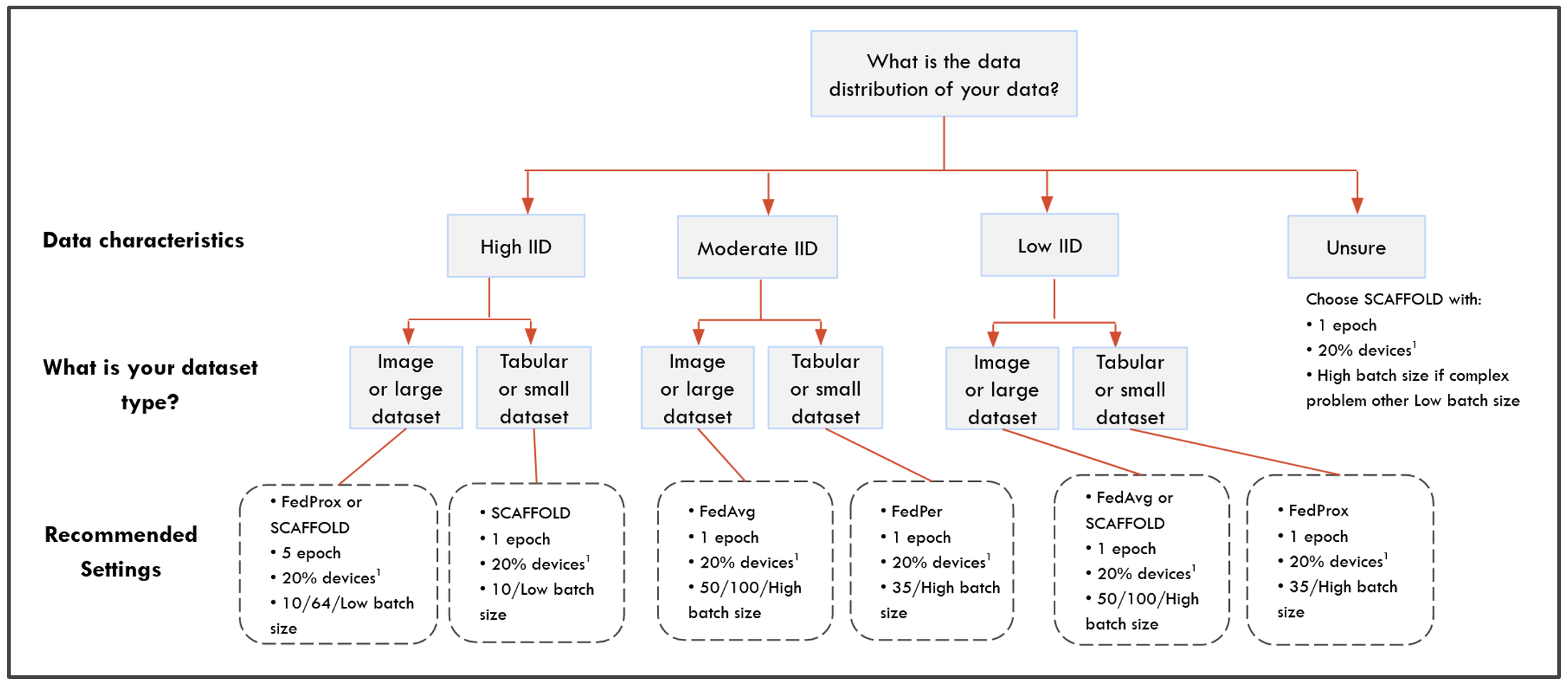}
\caption{Summary of Recommended Settings across IID Levels}
\label{fig:summary_recommended_settings}
\end{figure*}

In this study, we extensively tested four FL aggregators in three datasets and levels of IID. In each setting, we investigated three parameters: active devices, local epochs, and batch size, which impacts how devices learn and contribute to a global model in a federated setting. Our contribution is to further the understanding of how each aggregator operates under each level of statistical heterogeneity by each of the parameters, and further testing and validating the claims made by the initial authors of each of the aggregators. By parameter, our experiments indicated that using fewer active devices resulted in the best results regardless of IID setting. We also found that more epochs achieved better results in a high IID setting, but fewer epochs was preferable in a low IID setting, with mixed results in a moderate IID setting. Lastly, we found that best scores in a high IID setting were achieved with a smaller batch size, though in increasingly low IID settings, higher batch size was preferable. This was not true though of the CIFAR-10 dataset, which can be considered as a more complex task than the MNIST and Healthcare datasets.

Of the four FL aggregators, we observed that in the high IID setting no single aggregator regularly achieved the best F1 scores. FedAvg was consistently a strong performer in this setting, though not the best. Moving towards settings of greater statistical heterogeneity, FedAvg often achieved the best scores, which challenges some of the claims of the other aggregators that they would outperform FedAvg in low IID settings. SCAFFOLD emerged as a reliable alternative particularly in the low IID setting, particularly after removing outliers and volatility from CIFAR-10. Our study draws similar conclusions to Li et al. \cite{li2021federated}, who also found that state of the art aggregators did not regularly outperform FedAvg, though they cautioned against the use of SCAFFOLD for it is increased communication cost, since the size of the model parameters that need to be communicated are doubled through the control variates. 

As shown from the results, each aggregator has its strengths and weaknesses depending on the distribution of the dataset and the level of IID. This could be attributed to the nature of each aggregator and the way in which it handles statistical heterogeneity. Additionally, the performance of each aggregator depends on the specific characteristics and distribution of the dataset. The performance difference can also be affected by factors such as the data quantity variance and the robustness of the aggregator to changes in the parameter values, as described in the previous subsections. Based on the results from experimenting with different parameters and aggregators under high, medium and low levels of IID across MNIST, CIFAR-10 and Healthcare datasets, we are able to provide the following recommendations as indicative guidelines for future research studies of similar nature in Figure~\ref{fig:summary_recommended_settings}.

The scenario below demonstrates the process of determining the optimal parameters and settings to use based on the flow chart: the EMD heat maps shown in Figures 4 and 5 can be replicated by researchers to determine the EMD by the required number of class distributions and data quantity variance. Following this, they can classify the level of IID of their datasets of interest using the EMD score by referring to Table~\ref{table:emd_thresholds_values_k_labels}. Once the level of IID is determined, a researcher can then refer to the flow chart in Figure~\ref{fig:summary_recommended_settings} to select the optimal parameters that maximises model performance. For example, if an experiment requires using a 10-class, large sized dataset similar to MNIST and distributing 5 classes to the local devices, the IID level can be determined to be moderate, based on Table~\ref{table:emd_thresholds_values_k_labels}. Taking these into consideration and using the flow chart in Figure~\ref{fig:summary_recommended_settings}, we recommend using FedAvg, 1 epoch, 20\% of total local devices in global aggregation and a high batch size in the range of 50 to 100 to train the model. Rather than having to run multiple trials to find the suitable training condition that optimises model performance, the researcher can be benefited by having an evidence-based recommended setting as the starting point. 

\subsection{Limitations and Future Works}

We set out to explore the effect of data skewness on overall FL results by controlling label skew and data quantity skew. Through the preliminary experiments conducted to quantify overall skewness using EMD, it was brought to our attention that although label skew had a significant and consistent effect on the EMD metric, the data quantity skewness did not. Such results can be observed in the EMD heatmaps (Figures~\ref{fig:emd_analysis_mnist} and~\ref{fig:emd_analysis_healthcare}), with no distinguishable relationship between data quantity skewness and the EMD score. Therefore, we made a decision to only use label distribution as the determinator for data IID in our primary experiments. An explanation for the behaviour of the data quantity variance could exist in its relationship with the FL training methodology. One of the key features in our FL framework is that devices are selected at random at the beginning of each communication round, therefore this selection may potentially have a large impact on the overall degree of data skewness. For example, in an extremely low IID setting (for example, data quantity variance is 90\%), where a smaller proportion of active devices are selected (for instance, 10\% of all devices), it is possible that the selected devices do not all exhibit low IID data quantity settings, hence the training results may be similar or even exceed that of higher IID settings in any given communication round. 

An improvement to this measure could potentially explore limiting the data quantity variance within specified ranges for each mode of IID (for example, low IID falls within a range of 70\%-90\%). This would ensure that the active devices are all within their skewness range regardless of the random active device selection. Due to the exclusion of the data quantity variance in our experiments, a proposal for future work may be to explore the relationship between data quantity skewness and FL aggregator performance while keeping the label skew constant. These experiments would enable the complete analysis of the selected FL aggregators across different modes of skewness to determine their overall usability and effectiveness.

An evaluation limitation we found through our study was the lack of standard benchmarks for MNIST and CIFAR-10 FL models in the existing literature. Therefore, it was difficult to assess how our results compare to other combinations of parameters that have been reported. Related to this, we see future research benefitting from focussing on a wider range of datasets with different size and label imbalances, as we attempted to do by including the Healthcare dataset in our analysis. This would enable more robust validation of findings and generalisable conclusions to be formed. Another potential future direction involves the development of standardised benchmarks within the field of federated learning. This initiative aims to provide a consistent basis for future research, facilitating fair and meaningful comparisons of performance across different studies.
Lastly, most studies, including ours, optimise accuracy when evaluating model performance. An improvement to our study would be to repeat the experiments with multiple metrics in addition to accuracy, such as stability and communication efficiency, which could provide useful insights for a practical setting.


\section{Conclusion}

Our study has two main deliverables that we believe will add to the existing FL knowledge base. The first is our approach to determining the level of IID of a dataset, which has been extensively covered in Section~\ref{section:methods_datapartition}. The second is our recommended parameter settings for FL models under different levels of IID, which is also one of the research objectives for this study. We have achieved this by providing the flow chart in Figure~\ref{fig:summary_recommended_settings} to generalise our experimental results, which acts as a guidance for researchers who may face similar challenges in the future. 

Our proposed data partitioning approach tells which portion of the data is allocated to each device in a federated learning system, particularly to simulate statistical heterogeneity so that testing the robustness and performance of an aggregation algorithm can be tested better. Our proposed approach supports federated learning to ensure privacy concerns by keeping the raw data locally on each device, and only sharing model updates with the central server, thus preserving data privacy and adhering to the principles of federated learning.

It is important to note that these recommendations are subject to review and validation by further research, as our study did not perform exhaustive experiments on all types of dataset possible nor with a large variety of datasets in each category. Despite this, we believe that other researchers will find our preliminary results beneficial and informative when designing their research, especially when it comes to the selection of parameters and aggregators that optimise FL model performance.



\section*{Acknowledgment}
We would like to thank Maitreyee Satpathy, Rebecca Grace Johnston, and Jack Zi Jie Ye for their help and fruitful discussions for this study.

\bibliographystyle{unsrt}  
\bibliography{references}

\end{document}